\documentclass{article}
\usepackage{multibib}
\usepackage{arxiv}
\usepackage[utf8]{inputenc} 
\usepackage[T1]{fontenc}    
\usepackage{url}            
\usepackage{booktabs}       
\usepackage{amsfonts}       
\usepackage{nicefrac}       
\usepackage{microtype}      
\usepackage{lipsum}
\usepackage{amsmath}
\usepackage{amsthm}
\usepackage{amssymb}   
\usepackage{bigints}
\usepackage{bm}
\usepackage{diagbox}
\usepackage{makecell}
\usepackage{multirow}
\usepackage{caption}
\usepackage{subcaption}
\usepackage{graphicx}
\usepackage[ruled,vlined]{algorithm2e}
\include{pythonlisting}
\usepackage{xcolor}
\usepackage{mathtools}
\usepackage{enumitem}

\usepackage{hyperref}
\hypersetup{
    colorlinks=true,
    citecolor=black,
    linkcolor=black,
    filecolor=black,
    urlcolor=black,
}

\theoremstyle{definition}

\numberwithin{equation}{section}

\theoremstyle{plain}

\usepackage{array}
\newcolumntype{P}[1]{>{\centering\arraybackslash}p{#1}}

\usepackage{slashbox}

\title{History-Based, Bayesian, Closure for Stochastic Parameterization: Application to Lorenz '96}

\author{
  Mohamed Aziz Bhouri \\
  Department of Earth \\
  and Environmental Engineering \\
  Columbia University\\
  New York, NY 10027 \\
  \texttt{mb4957@columbia.edu } \\
  \And
  Pierre Gentine \\
  Department of Earth \\
  and Environmental Engineering \\
  Department of Earth \\
and Environmental Sciences \\
  Columbia University\\
  New York, NY 10027 \\
  \texttt{pg2328@columbia.edu} \\
}

\begin{document}
\maketitle

\begin{abstract}

Physical parameterizations are used as representations of unresolved subgrid processes within weather and global climate models or coarse-scale turbulent models, whose resolutions are too coarse to resolve small-scale processes. These parameterizations are typically grounded on physically-based, yet empirical, representations of the underlying small-scale processes. Machine learning-based parameterizations have recently been proposed as an alternative and have shown great promises to reduce uncertainties associated with small-scale processes. Yet, those approaches still show some important mismatches that are often attributed to stochasticity in the considered process. This stochasticity can be due to noisy data, unresolved variables or simply to the inherent chaotic nature of the process. To address these issues, we develop a new type of parameterization (closure) which is based on a Bayesian formalism for neural networks, to account for uncertainty quantification, and includes memory, to account for the non-instantaneous response of the closure. To overcome the curse of dimensionality of Bayesian techniques in high-dimensional spaces, the Bayesian strategy is based on a Hamiltonian Monte Carlo Markov Chain sampling strategy that takes advantage of the likelihood function and kinetic energy's gradients with respect to the parameters to accelerate the sampling process. We apply the proposed Bayesian history-based parameterization to the Lorenz '96 model in the presence of noisy and sparse data, similar to satellite observations, and show its capacity to predict skillful forecasts of the resolved variables while returning trustworthy uncertainty quantifications for different sources of error. This approach paves the way for the use of Bayesian approaches for closure problems.

\end{abstract}


\section*{Lead Paragraph}

\textbf{Climate models involve physical processes with different scales. Given the available computational resources, small-scale processes are not resolved in global climate models but rather represented by parameterization schemes. Machine learning has been recently used to improve existing parameterization approaches, yet those methods still show some important mismatches that are often attributed to stochasticity in the considered processes. This stochasticity can be due to noisy data, unresolved physical variables or simply to the inherent chaotic nature of the process. In this work, we develop a probabilistic parameterization scheme capable of predicting skillful forecasts of the resolved physical variables while returning trustworthy uncertainty quantifications for different sources of error. This approach paves the way for the use of Bayesian approaches for parameterization problems.}

\section{Introduction}

Parameterization schemes, or closures, are approximate representation of unresolved subgrid processes in weather and climate models and are the most dominant source of uncertainty in models predictions. The corresponding errors have been reduced throughout improvements of existing parameterizations and the development of new schemes, yet these errors cannot be completely eliminated because of inherent model structural errors, as they try to approximate complex physical processes. To address these structural errors, recently, several groups have started developing machine-learning based parameterizations, which have been shown to dramatically improve the representation of physical processes and strongly reduce parameterization structural errors compared to standard parameterizations \cite{Brenowitz2018,Rasp2018,Gentine2018,OGorman2018,Bolton2019}.
Another source of uncertainty stems from the inherent stochastic nature of many physical processes in nature \cite{Lorenz96,Palmer2001,Tribbia2004,Karimi2010}. In order to better characterize this latter source of uncertainty and  quantify the corresponding errors, stochastic parameterization schemes have been proposed in order to approximate the probability distribution of  subgrid scale processes based on the coarse-scale resolved variables. Such approximations are generally built as an ensemble of parameterization estimates which results in an ensemble prediction system \cite{Palmer2012}.

As an example, stochastic parameterizations for weather prediction models have been shown to significantly improve the forecasts quality across different timescales including at sub-seasonal and seasonal time-frames and are now extensively used by meteorological weather forecasting groups around the world \cite{palmer2008a,Reyes2009,Berner2009,Stockdale2011,Palmer2012,Weisheimer2014,Sanchez2016,Leutbecher2017}. Stochastic modeling has also gained importance in climate prediction and projection, and several studies have demonstrated the improvements gained through adopting stochastic parameterization schemes. These improvements consist of, but are not limited to, (1) more faithful long-time integration of climate models \cite{Wang2016,Davini2017,Christensen2017,Strommen2018}, (2) more accurate model mean state estimates \cite{Berner2012}, (3) better characterization of climate variability and sensitivity \cite{Seiffert2010,Ajayamohan2013,Dawson2015} and (4) more stable dynamical systems \cite{Berner2017}. These advances have narrowed the gap between weather and climate predictions since global cloud-resolving models are capable of forecasting weather as well as providing faithful longer-term (up to a year or 2) climate predictions \cite{Crueger2018,Satoh2019} and weather forecasting centers predictions are not limited to weekly time scales but extend to seasonal and multi-year time frames \cite{Moncrieff2007,Vitart2012}. The interconnection between weather and climate predictions stems from the nonlinear energy transfer from smaller to larger scales \cite{Lorenz96,Palmer2001,Tribbia2004} and from the predictable signals on the higher frequencies that are produced by the slower variability modes \cite{Hoskins2013,Vannitsem2016}. Parameterization schemes are a critical part of this connection between weather and climate.

Different approaches have been proposed for parameterization schemes. We can distinguish between simple approaches which are extensively used thanks to their simple implementation and model prediction improvements \cite{Buizza1999,Sanchez2016,Christensen2017,Leutbecher2017} and statistical schemes such as for convection and cloud formation \cite{Craig2006,Khouider2010,Vissio2018a,Vissio2018b,Sakradzija2018,Bengtsson2019,Gutierrez2021}. Data-driven approaches are another option to build parameterizations based on observations or high-resolution model simulations in order to account for the variability that is unresolved in the lower-resolution climate models \cite{Shutts2007,Shutts2014,Dorrestijn2015,Christensen2015b,bessac2019a,Christensen2020}. The major limitation of these approaches is that they assume that all required variables are accessible to the machine-learning model, whereas instantaneous observations are typically only available for coarse-scale variables, and these techniques assume a deterministic relationship, as most standard machine learning methods are deterministic. Thus, these methods might miss important internal variability which might be important for prediction. 

Machine learning-based methods have been successfully developed in order to parameterize various atmospheric \cite{Krasnopolsky2005,Schneider2017,Brenowitz2018,Rasp2018,Gentine2018,OGorman2018} and oceanic processes \cite{Bolton2019}. Although non-deterministic methods, e.g. Generative Adversarial Networks, have recently been used to parametrized chaotic Lorenz '96 system \cite{Gagne2020}, most of machine learning-based parameterizations are deterministic approaches. Bayesian surrogate models allow for uncertainty quantification and several inference techniques have been recently proposed in order to generalize deep learning methods to Bayesian frameworks. These Bayesian deep learning techniques have been successfully employed in several fields such as language processing \cite{Xiao2019}, computer vision \cite{Kendall2017}, differential equations solutions estimate \cite{Yang2020,Bhouri2021}, sequential decision making \cite{Riquelme2018} and black-box function optimization \cite{Snoek2015}.

Bayesian inference techniques include ensemble methods which consist in training the same machine learning surrogate model but with different initializations in order to capture the model's epistemic (internal) uncertainty \cite{Lakshmin2017,Osband2018,Fort2019,Ciosek2020}. The major advantage of the ensemble methods compared to sampling methods is their capability to better characterize multi-modal posterior distributions when minimizing non-convex loss functions with several local minima. This behavior has been shown in the context of image classifications, for instance \cite{Fort2019}. Within the same concept, the Stochastic Weight Averaging (SWA) method takes advantage of the many local minima of non-convex loss functions by considering the average of multiple neural network parameters along the training process \cite{Izmailov2018}. The SWA employs a cyclical learning rate in order to enhance the stochastic gradient descent ability to further explore the non-convex loss manifold, which favors finding the local minima. Dropout (i.e. dropping randomly connections between neurones) provides another alternative to build an ensemble of neural networks given a pre-chosen probability \cite{Srivastava2014,Gal2016}. Hence, dropout is able to provide some uncertainty quantification in neural networks estimates with a low computational cost. However, it has been shown that dropout often  strongly underestimates uncertainty estimates, which prevents its use in applications requiring sufficiently accurate approximation of the full uncertainty via predicted posterior distributions \cite{Riquelme2018,Osband2022}. Variational inference is another strategy for Bayesian inference and relies on approximating typically intractable posterior distributions with tractable ones \cite{Hinton1993,Blundell2015}. This approximation enables using stochastic gradient descent methods in order to update the model's parameters (weights and biases) since the objective function is tractable. However, these approximations result in posterior variance underestimation and when variational inference is applied to deep learning frameworks it results in a poor approximation of the true multi-modal posterior distribution \cite{Osband2022}.

Although Markov-Chain Monte Carlo (MCMC) sampling approaches are considered as the gold-standard for Bayesian inference \cite{Cowles1996,Jespersen2010,Roy2020}, their high computational cost severely limits their application to large-scale problems and deep neural networks \cite{Ravenzwaaij2018}. However, Hamiltonian Monte Carlo (HMC) methods \cite{Neal2011} offer a viable Markov-Chain sampling alternative. Similarly to Hamiltonian mechanics principles, these methods are based on characterizing the parameters sampling process as a motion governed by a Hamiltonian function which is the sum of potential and kinetic energy. Such a formalism allows defining a dynamical system governing the neural network parameters' evolution and their time-step integration is based on a Markov Chain that approximates the posterior distribution of the inferred parameters \cite{Betancourt2017}.
The Hamiltonian dynamical system depends on the gradient of the likelihood function and kinetic energy with respect to the parameters. Therefore, the gradient information can be used to accelerate the convergence of the MCMC sampling process while searching for the most likely posterior distribution for the parameter space of the neural networks \cite{Yang2020}. 

All these Bayesian inference techniques offer a new possibility to build fully-Bayesian parameterization schemes that are capable of returning state-dependent uncertainty quantification, which cannot be obtained by deterministic parameterization schemes. We here present the first application of HMC methods for parameterizations in the Earth sciences.

Besides, existing parameterization methods rely on a parameterized model that depends mostly on the current model time step. However, many physical processes, such as turbulence or clouds, have substantial memory. Excluding memory from the parameterization (closure) can lead to further uncertainties and appear as another source of stochasticity, as the model will generate erroneous predictions given only the current time state. Such modeling choice might limit considerably the online forecasts of the parameterized model. This is due to the fact that subsequent states of resolved variables cannot be determined based only on current state of these variables but also on the state of unresolved variables.

To overcome these major limitations, we develop a history-based parameterization that relies not only on the current state of the resolved variables, but also on their previous states. We show that the time integration of the parameterized dynamical system returns trajectories of the resolved variables that are closer to the true state of the state, even when evaluated online (i.e. coupled within the dynamical system). The proposed parameterization is only trained using data of the resolved variables. We show that the parameterization is capable of not only learning the coupling terms but also to account and correct for the numerical error introduced by the temporal discretization, which enhances stability and accuracy of the parameterized system's resolution. The history-based parameterization is then extended into a stochastic formulation by relying on the Bayesian formalism based on the HMC Markov Chain sampling. We apply the proposed history-based stochastic parameterization to the Lorenz '96 model with noisy and sparse simulated observational data and show its capability of producing accurate temporal forecasts for the resolved variables while returning faithful uncertainty quantification for the different sources of error.

The Lorenz '96 model  and the proposed history-based parameterization scheme are detailed in section \ref{sec:methods}. For clarity, we distinguish between the deterministic and Bayesian history-based parameterization schemes. The forecast results and study of numerical error and stability of the proposed parameterization schemes are presented in section \ref{sec:res}. The proposed history-based parameterization schemes are evaluated against deterministic and Bayesian non-history-based parameterization schemes. We also detail the uncertainty quantification obtained with each parameterization scheme for different sources. Finally, in section \ref{sec:cl}, we summarize the proposed parameterization schemes and their results, discuss shortcoming of the proposed approach and carve out directions for future investigation.

\section{Methods}
\label{sec:methods}

\subsection{Lorenz '96 Model} 
\label{subsec:L96}

The Lorenz '96 model is a two-time scale dynamical system which mimics the non-linear dynamics of the extratropical atmosphere with simplified representation of multiscale interactions and nonlinear advection \cite{Lorenz96}. It consists of a set of equations coupling variables evolving over slow and fast timescales:
\begin{align}
    \label{eq:L96_X}\frac{d X_k}{dt} = -X_{k-1}(X_{k-2}-X_{k+1})-X_k+F-\frac{hc}{b}\sum\limits_{j=J(k-1)+1}^{kJ}Y_{j} \ ,k=1,\ldots,K \\
    \label{eq:L96_Y}\frac{d Y_{j}}{dt}=-cbY_{j+1}(Y_{j+2}-Y_{j-1})-cY_{j}+\frac{hc}{b}X_{\lfloor(j-1)/J\rfloor+1} \ , j=1,\ldots,JK \ .
\end{align}

The model includes $K$ large-scale, low-frequency (slow-varying) variables $X_k$, $k=1,\ldots,K$. Each slow-varying variable $X_k$ is coupled to a larger number of small-scale, high-frequency (fast-varying) variables $Y_j$, $j=J(k-1)+1,\ldots k J$. The fast time scales impact the slow variables through the coupling term $\sum_{j=J(k-1)+1}^{kJ}Y_{j}$ and the coupling strength depends on three key parameters: $b$, $c$ and $h$. The parameter $b$ determines the magnitude of the non-linear interactions between the fast variables. The parameter $c$ controls how rapidly the fast-varying variables fluctuate compared to the slow-varying variables. Finally, the parameter $h$ governs the strength of the coupling between the slow- and fast-varying variables.

The chaotic dynamical system Lorenz '96 is very useful for testing different numerical methods in atmospheric modeling thanks to its transparency, low computational cost and simplicity compared to full-blown Global Climate Models (GCM). The interaction between variables of different scales makes the Lorenz '96 model of particular interest when evaluating new parameterization methodologies. As such, it was used in assessing different techniques that were later incorporated into GCMs \cite{Crommelin2008,Dorrestijn2013}.

The Lorenz '96 model has been extensively used and studied as a test bed in various studies including data assimilation approaches \cite{law2016,Hatfield2018}, stochastic parameterization schemes \cite{Kwasniok2012,Arnold2013,Chorin2015} and machine learning-based parameterizations  \cite{Schneider2017,Dueben2018, Watson2019,Gagne2020}. Finally, we note that the forcing term $F$ in equation (\ref{eq:L96_X}) was missing in the original Lorenz '96 paper \cite{Lorenz96}, but the author provides the parameter. All other authors have then used this forcing term as well (e.g. \cite{Wilks2005}). The full Lorenz '96 system (\ref{eq:L96_X})-(\ref{eq:L96_Y}) including the fast variables is considered as the ``true'' model and is used to generate the dataset.  

\subsection{Standard Parameterizations}
\label{subsec:std_param}

The instantaneous parameterizations for the Lorenz '96 model replace the coupling term as a function of the slow-varying variables $X_k$, $k=1,\ldots K$ at the current time step:
\begin{equation}
    \label{eq:L96_param_other}\frac{d X_k^*}{dt} = -X^*_{k-1}(X^*_{k-2}-X^*_{k+1})-X^*_k+F+P(X_k^*;\bm{\theta}) \ ,k=1,\ldots,K ,
\end{equation}
\noindent where $X^*_k$, $k=1,\ldots,K$ is the forecast estimate of $X_k$ based on the parameterized subgrid tendency $P(\cdot;\cdot)$ and $\bm{\theta}$ is a vector of unknown parameters that are learned given the available dataset.

In our study, the goal is to infer a surrogate model for the parameterization (closure) that replaces the coupling term. We also aim to build a parameterization that remains accurate over long time periods when tested online. As we will show, this task is typically unfeasible unless we allow the parameterized subgrid tendency term to depend not only on the slow-varying variables evaluated at the current time, but also on the previous time steps. Indeed, relying on a parameterization of the form in equation (\ref{eq:L96_param_other}) will likely not provide an online forecast estimate $X_k^*$, $k=1,\ldots,K$, that ``matches'' the ``true'' model variables $X_k$, $k=1,\ldots,K$, since the state of these variables at a time instance $t$ does not only depend on their state at a previous time instance, but also on the state of the fast-varying variables $Y_j$, $j=1,\ldots JK$, at the same previous time instance. In other words, in a forecast setting, the next time-step prediction for $X_k^*$, $k=1,\ldots,K$, does not only depend on its current time state but also on the current state of the inaccessible small-scale variables $Y_j$, $j=1,\ldots JK$.

\subsection{History-Based Parameterization} 
\label{sec:hist_param} 

The goal of the proposed method is to only use the slow-varying variables in order build a parameterization, i.e. closure, that replaces the fast-to-slow variables' coupling terms, along with uncertainty quantification. Such a parameterization should result in a dynamical system that only depends on the slow-varying variables and whose time integration provides estimates that are sufficiently close to the states obtained with the ``true'' model. Hence, we target a parameterization that is accurate when tested online, i.e. when integrating (in time) the parameterized system based only on the slow variables. This goal is motivated by the constraints and requirements observed for weather and climate predictions. For instance, in order to model the atmosphere, it is not feasible to explicitly simulate the smallest scales due to computational limits, even though the physical equations of motions are known. This is due to our limited computational resources, and hence parameterization schemes are needed to represent small scale processes and solve this issue. In the case of the Lorenz '96 system, a forecast model is built by truncating the variables to only the slow-varying ones $X_k$, $k=1,\ldots K$, and by parameterizing the impact of the fast-varying variables $Y_j$, $j=1,\ldots JK$ on the resolved ones $X_k$, $k=1,\ldots K$. 

\subsubsection{Deterministic Formulation}
\label{sec:det_param} 

To remedy the issues stemming from the inference based on instantaneous parameterizations, we propose a parameterized subgrid tendency that depends not only on the slow-varying variables evaluated at the current time, but also on their states at previous time-steps as follows:
\begin{equation}
    \label{eq:L96_param_gen}\frac{d X_k^*}{dt} = -X^*_{k-1}(X^*_{k-2}-X^*_{k+1})-X^*_k+F+P\big(X_k^*(t),X_k^*(t-\tau_1),\ldots,X_k^*(t-\tau_{n_h});\bm{\theta}\big) \ ,k=1,\ldots,K ,
\end{equation}
\noindent where $\tau_i$, $i=1,\ldots n_h$ are the time lags that define the number of previous time steps being considered for the parameterization. With such a parameterization, we want to assess whether the effect of the fast-varying variables on the forecast of the slow-varying variables is (partly) embedded within the historical (previous time-steps) evolution of the slow-varying variables. Such history-based inference can be carried out using machine learning techniques. 

In the current work, and as it is for atmosphere dynamics parameterization, we assume that we have access to a discretized (and noisy, see below) time trajectory of the slow-varying variables of the ``true'' model. Therefore, relying on a parameterized subgrid closure that depends not only on the slow-varying variables evaluated at the current time, but also at previous time-steps does not add any constraint or requirement regarding the data that is needed to construct the proposed parameterization. Besides, the time-step $\Delta t$ of the available data can be used to define the time lags $\tau_i$, $i=1,\ldots n_h$ in (\ref{eq:L96_param_gen}) as detailed below. 

One challenge that may result from considering a history-based parameterization is that the resulting dynamical system (\ref{eq:L96_param_gen}) will consist of a Delay Differential Equation (DDE) instead of an Ordinary Differential Equation (ODE). Using Runge-Kutta schemes for DDEs may result in interpolating the existing data in order to perform the time-marching \cite{Enright1995,Fudziah2002}, which can be an additional significant source of error for the inference task. Fourth-order Runge-Kutta (RK4) schemes are the standard time stepping methods that are used not only to solve the ``true'' model but also to solve the parameterized one \cite{Lorenz96,Wilks2005,iserles2008,Arnold2013,Gagne2020}. Explicit RK4 schemes are suitable for such a parameterization problem given their stability properties, and the fact that using implicit schemes would result in a computational bottleneck when using machine learning surrogate models for the parameterization. In such a case, performing time-marching with an implicit time stepping scheme would require solving non-linear equations depending on the output of the machine learning surrogate models that are used for the parameterization, which is generally infeasible when using deep neural networks. 

We note that the use of time memory leads to some technical challenges in the numerical integration of the model, that we detail in Appendix  \ref{app_sec:hist_param}. As a summary, fitting discrete observations of the slow-varying variables that are available with a time-step $\Delta t$ by discretizing the DDE (\ref{eq:L96_param_gen}) with an explicit RK4 can be carried out without interpolating data by choosing the time-lags as multiples of $2 \ \Delta t$: $\tau_i = 2 i \Delta t$ , $i=1,\ldots,n_h$ and discretizing the DDE (\ref{eq:L96_param_gen}) with the time-step $2 \ \Delta t$. We also detail the differentiable time solver implemented to learn the parameterization (\ref{eq:L96_param_gen}) in Appendix \ref{app_sec:hist_param}. A system identification task can be formulated as follows. Given some observations $\mathcal{D}=\{X(t_i) , t_i = i\Delta t , i=0,\ldots,n-1\}$, we can learn the vector $\bm{\theta}$ that best parameterizes the underlying dynamics (\ref{eq:L96_param_gen}) by defining a loss function $\mathcal{L}$ that measures the discrepancy between the observed data and the parameterized model predictions for a given $\bm{\theta}$. If the dynamical system (\ref{eq:L96_param_gen}) is integrated $n_f$ times for each data-point considered, then the loss can be formulated as follows:
\begin{equation}
    \label{eq:loss_batch}\mathcal{J}(\bm{\theta};n_f) = \frac{1}{n_b}\sum_{j\in B} \mathcal{L}\Big(\bm{X}(t_{j+n_f+1}),\bm{X}^*(t_j+ (n_f+1) \Delta t;\bm{\theta})\Big)\ ,
\end{equation}
\noindent where $B$ is a random subset of $\{0,\ldots,n-n_f-2\}$ of size $n_b$. $X^*(t_j+(n_f+1) \Delta t;\bm{\theta})$ is the parameterized prediction obtained for the data-point $\bm{X}(t_{j+n_f+1})$ by integrating (\ref{eq:L96_param_gen}) $n_f$ times and starting from the data-point $\bm{X}(t_j)$ , $j=0,\ldots,n-n_f-2$. 

Numerical integration (e.g. doubling the time step) of the parameterized model may be a source of additional numerical error. However, as we will show in the numerical results, section \ref{sec:res}, the machine learning surrogate model $P$ used in equation (\ref{eq:L96_param_gen}) is actually not only capable of learning the coupling term, but also to correct for the numerical errors introduced by the discretization of the parameterized model compared to the accuracy of the data on which the model is trained. Indeed, we show that the parameterized model (\ref{eq:L96_param_gen}) discretized with a time-step equal to $2 \ \Delta t$ and trained on a data generated with the ``true'' model (\ref{eq:L96_X})-(\ref{eq:L96_Y}) which was solved with a finger time-step equal to $\Delta t/2$, can generate forecasts that closely match the ``true'' model trajectories, while using the RK4 to solve the ``true'' model (\ref{eq:L96_X})-(\ref{eq:L96_Y}) with a time-step equal to $2 \ \Delta t$ is not even computationally stable. This means that the parameterization is not only able to correct the numerical errors, but also to ensure the numerical stability of the parameterized model even when the latter is discretized with a larger time step (time step equal to $2 \ \Delta t$) for training and forecasting. The deterministic inference approach can also be applied for standard instantaneous parameterizations described in section \ref{subsec:std_param} by discretizing the parameterized dynamical system (\ref{eq:L96_param_other}) with a time-step $\Delta t$.

The framework detailed so far is only capable of providing deterministic estimates for machine learning surrogate model's parameters $\bm{\theta}$, which correspond to a (local) minimum of the deterministic loss $\mathcal{J}$ (\ref{eq:loss_batch}). However, for the Lorenz '96 parameterization, as it is the case for atmospheric predictions, several sources of uncertainty should be taken into consideration. These sources of uncertainty consist of (1) the chaotic and/or stochastic nature of the dynamical system, (2) the limitation introduced by the parameterization with a lower set of predictors compared to the ``true'' model (which includes fast variables), (3) the quality of the data used to retrieve the closure, given potential noise and temporal sparsity in those observations and (4) the discretization of the continuous differential equations. Therefore, it is desirable to obtain a distribution over plausible machine learning surrogate model's parameters that can effectively characterize these different sources of uncertainty in the estimates.

\subsubsection{Bayesian Formulation with Hamiltonian Monte Carlo} 
\label{sec:bayes_param}

Quantifying uncertainties can be carried out by using a Bayesian formalism, with prior information that is available for the unknown parameters $\bm{\theta}$ and the quality of the observations in the training dataset, in terms of temporal sparsity and noise as detailed in the next sections.

Bayesian inference provides a statistical characterization of the posterior distribution of the inferred parameters $\bm{\theta}$,  that can be formulated as follows:
\begin{multline}
    \label{eq:post}p(\bm{\theta},\gamma,\lambda|\bm{X}(t+2 \ \Delta t), \bm{X}(t),\ldots, \bm{X}(t-(2n_h-1)\Delta t), \bm{X}(t-2n_h\Delta t))\propto  \\
    p(\bm{X}(t+2 \ \Delta t)| \bm{X}(t),\ldots, \bm{X}(t-(2n_h-1)\Delta t), \bm{X}(t-2n_h\Delta t),\bm{\theta},\gamma)p(\bm{\theta},\lambda)p(\lambda)p(\gamma) \ , 
\end{multline}
\noindent where $p(\bm{X}(t+2 \ \Delta t)| \bm{X}(t),\ldots, \bm{X}(t-(2n_h-1)\Delta t), \bm{X}(t-2n_h\Delta t),\bm{\theta},\gamma)$ is a likelihood function that measures the discrepancy between the observed data and the parameterized model's predictions obtained with the time-stepping scheme described in section \ref{sec:det_param} for given parameters $\bm{\theta}$, $p(\bm{\theta}|\lambda)$ is a prior distribution parameterized by the variable $\lambda$ that encodes any prior domain knowledge about the unknown model parameters $\bm{\theta}$, and $\gamma$ characterizes the quality of the available observations in terms of noise and temporal sparsity for instance. The likelihood and priors are chosen based on a hierarchical Bayesian approach \cite{Yang2020} as follows:
\begin{multline}
    \label{eq:likelihood}p(\bm{X}(t+2 \ \Delta t)| \bm{X}(t),\ldots, \bm{X}(t-(2n_h-1)\Delta t), \bm{X}(t-2n_h\Delta t),\bm{\theta},\gamma)= \\
    \prod_{i=0}^{2n-2}\mathcal{N}\Big(\bm{X}(t_i+2\ \Delta t)|\mathcal{F}\big(\bm{X}(t_i),\ldots, \bm{X}(t_i-(2n_h-1)\Delta t), \bm{X}(t_i-2n_h\Delta t),\bm{\theta}\big),\gamma^{-1}\Big) \ ,
\end{multline}
\begin{align}
    \label{eq:prior_theta}p(\bm{\theta}|\lambda) = \mathrm{Laplace}(\bm{\theta}|0,\lambda^{-1}) \ , \\
    \label{eq:prior_lambda}p(\log \lambda) = \mathrm{Gam}(\log \lambda|\alpha_1,\beta_1) \ , \\
    \label{eq:prior_gamma}p(\log \gamma) = \mathrm{Gam}(\log \gamma|0,\alpha_2,\beta_2) \ ,
\end{align}
\noindent where $\mathcal{F}(\cdot)$ denotes the time-stepping scheme to integrate the parameterized DDE (\ref{eq:L96_param_gen}) and detailed in Appendix \ref{app_sec:hist_param}.

We choose a Gaussian likelihood function since we assume an uncorrelated Gaussian noise with zero mean and a precision $\gamma$ that is inferred from the observed data and the model predictions through Markov Chain sampling as detailed below. The Gamma distribution is a common choice as prior for the unknown precision parameters $\gamma$ and $\lambda$, but other alternatives exist \cite{winkler1967,bernardo1979,berger1990,geweke1993,gelman2013}. The logarithm of the precision parameters $\gamma$ and $\lambda$ is used to ensure positive values for these variables during the training process. The Laplace prior for the model parameters $\bm{\theta}$ is chosen in order to promote sparsity within the inferred parameters which helps avoid overfitting since deep neural networks will be used as surrogate models for the parameterization. Using a Laplace prior for the neural network parameters acts as a regularizer by promoting sparsity among the inferred parameters, similarly to using an $L_1$ loss within a deterministic setting.

The posterior distribution (\ref{eq:post}) cannot usually be derived analytically, given the modeling assumptions for the likelihood and prior distributions (\ref{eq:likelihood})-(\ref{eq:prior_gamma}), and the nonlinearity of the dynamical system (\ref{eq:L96_param_gen}), so that the resulting distribution will not be Gaussian. Sampling from this unnormalized non-Gaussian distribution in order to infer it is challenging and computationally expensive, mainly when the model parameters $\bm{theta}$ are relatively high-dimensional, which is the case when using deep neural networks as surrogate models for the parameterization. For such an application the parameter space dimension is typically in the order of million(s) \cite{Rasp2018}. Given this constraint, we here use a Hamiltonian Monte Carlo (HMC) technique \cite{Neal2011} since it is a powerful method to tackle Bayesian inference tasks in high dimensions. HMC allows using gradient of the likelihood function and kinetic energy with respect to the parameters, as detailed below, to effectively generate approximate posterior samples for the inferred parameters. This property allows computational speedup by relying on graphical processing units (GPUs) to efficiently back-propagate the gradient, as it is the case for generic machine learning training processes.

Let $\bm{\Theta}=[\bm{\theta},\gamma,\lambda]$ denote the vector of all unknown parameters to be inferred from the available dataset using HMC. Then, we define a Hamiltonian function, by analogy with Hamiltonian mechanics formalism, $H=U(\bm{\Theta})+V(\bm{v})$, where $U(\bm{\Theta})$ is the potential energy of the original system, which is defined as the logarithm of the likelihood function (\ref{eq:likelihood}), and $V(\bm{v})$ is the kinetic energy of the system depending on the ``velocity'' defined as $\bm{v}=d\bm{\Theta}/dt$, where $dt$ is the increment of the HMC algorithm. The evolution of $\bm{\Theta}$ and $\bm{v}$ is then given by the following equations depending on the gradients of the Hamiltonian:
\begin{equation}
    \label{eq:HMC_ODE}\frac{d\bm{v}}{dt} = -\frac{\partial H}{\partial \bm{\Theta}} \ , \ \frac{d\bm{\Theta}}{dt}=\frac{\partial H}{\partial \bm{v}} .
\end{equation}

Integrating this dynamical system (\ref{eq:HMC_ODE}) in time, using an energy-preserving leapfrog scheme \cite{Neal2011}, provides a Markov chain that converges to the (stationary) posterior distribution (\ref{eq:post}). Accurate and efficient computation of the gradient $\partial H/\partial \bm{\Theta}$ is made possible using a differentiable time-solver that is detailed in Appendix \ref{app_sec:hist_param} and needed to evaluate the likelihood function (\ref{eq:likelihood}). Gradient evaluations can then be carried out sufficiently fast by relying on GPUs to effectively back-propagate  the gradient through the time-solver. Surrogate model parameters $\bm{\theta}$ and precision parameters $[\gamma,\lambda]$ can be update either simultaneously or separately using Metropolis-within-Gibbs schemes \cite{gilks1995,millar2000}, we here update them simultaneously.

\subsubsection{Generating Forecasts with Quantified Uncertainty}

The HMC method returns a faithful set of parameters samples that concentrate in regions of high probability in the posterior distribution space of $p(\bm{\theta},\gamma,\lambda|\mathcal{D})$. These samples allows the inference of future forecasts for $\bm{X}^*$ in time with quantified uncertainty by computing the following predictive posterior distribution
\begin{multline}
    \label{eq:post_pred}p(\bm{X}^*(t)|\mathcal{D},\bm{X}(0),\bm{X}(\Delta t),\ldots,\bm{X}((2n_h+1)\Delta t),t)= \\ \int p(\bm{X}^*(t)|\bm{\Theta},\bm{X}(0),\bm{X}(\Delta t),\ldots,\bm{X}((2n_h+1)\Delta t),t) p(\bm{\Theta}|\mathcal{D}) \mathrm{d}\bm{\Theta} \ .
\end{multline}

This predictive posterior distribution provides a complete statistical characterization for the forecasted states by accounting for the epistemic uncertainty in the inferred dynamics, and for the quality of the available data (in terms of noise and temporal discretization and sparsity for instance) on which the model was trained. Data sparsity is modeled in this work by integrating the ``true'' model (\ref{eq:L96_X})-(\ref{eq:L96_Y}) using a small time-step and using a subsample of the resolved variables trajectories in order to train the parameterized model on observations spaced with a larger time-step as detailed in section \ref{sec:pb_setup}. Noise is accounted for by corrupting the resolved variables observations with a Gaussian noise whose standard deviation is proportional to the standard deviation of the resolved variables trajectories as detailed in section \ref{sec:pb_setup}. Plausible realizations of $\bm{X}^*(t)$ can be sampled from this predictive distribution as follows:
\begin{equation}
    \label{eq:real_X}\bm{X}^*(t)= \mathcal{G}(\bm{X}(0),\bm{X}(\Delta t),\ldots,\bm{X}((2n_h+1)\Delta t),t,\bm{\theta})+\epsilon \ , \ \epsilon\sim\mathcal{N}(0,\gamma^{-1}) \ , \ \bm{\theta},\gamma,\lambda\sim p(\bm{\theta},\gamma,\lambda| \mathcal{D}) \ ,
\end{equation}
\noindent where $\mathcal{G}(\cdot)$ denotes the operation of performing the time-stepping scheme $\mathcal{F}(\cdot)$ multiple times in order to predict $\bm{X}^*(t)$ from the initial conditions $\bm{X}(0),\bm{X}(\Delta t),\ldots,\bm{X}((2n_h+1)\Delta t)$ needed to solve the DDE (\ref{eq:L96_param_gen}), and $\bm{\theta},\gamma,\lambda$ are approximate samples from the posterior distribution $p(\bm{\theta},\gamma,\lambda|\mathcal{D})$ computed during model training via HMC. $\epsilon$ accounts for the noise that may be corrupting the observations that are used for training.

To understand the behavior of the probabilistic trajectories, maximum \textit{a posteriori} (MAP) estimate of the surrogate model (neural network) and precision parameters is considered:
\begin{equation}
    \label{eq:MAP_param} \bm{\theta}_{\mathrm{MAP}},\gamma_{\mathrm{MAP}},\lambda_{\mathrm{MAP}} = \mathrm{arg}\max_{\bm{\theta},\gamma,\lambda}p(\bm{\theta},\gamma,\lambda| \mathcal{D}) \ .
\end{equation}

These MAP surrogate model and precision parameters are used to obtain the MAP states and model trajectories $\bm{X}^*_{\mathrm{MAP}}$ as follows:
\begin{equation}
    \label{eq:MAP_X}\bm{X}^*_{\mathrm{MAP}}(t)= \mathcal{G}(\bm{X}(0),\bm{X}(\Delta t),\ldots,\bm{X}((2n_h+1)\Delta t),t,\bm{\theta}_{\mathrm{MAP}})+\epsilon \ , \ \epsilon\sim\mathcal{N}(0,\gamma_{\mathrm{MAP}}^{-1}) \ .
\end{equation}

The posterior parameters samples can also provide an approximation for the first- (mean) and second-order (variance) statistics of the predicted states $\bm{X}^*$ as:
\begin{multline}
    \label{eq:mu_X}\mu_{\bm{X}^*}(t)= \int\mathcal{G}(\bm{X}(0),\bm{X}(\Delta t),\ldots,\bm{X}((2n_h+1)\Delta t),t,\bm{\theta})p(\bm{\theta}|\mathcal{D})\mathrm{d}\bm{\theta} \\
    \approx\frac{1}{N_s}\sum_{i=1}^{N_s} \mathcal{G}(\bm{X}(0),\bm{X}(\Delta t),\ldots,\bm{X}((2n_h+1)\Delta t),t,\bm{\theta}_i)\ ,
\end{multline}
\begin{multline}
    \label{eq:sigma_X}\sigma^2_{\bm{X}^*}(t)= \int\Big(\mathcal{G}(\bm{X}(0),\bm{X}(\Delta t),\ldots,\bm{X}((2n_h+1)\Delta t),t,\bm{\theta})-\mu_{\bm{X}^*}(t)\Big)^2 p(\bm{\theta}|\mathcal{D})\mathrm{d}\bm{\theta} \\
    \approx\frac{1}{N_s}\sum_{i=1}^{N_s} 
    \Big(\mathcal{G}(\bm{X}(0),\bm{X}(\Delta t),\ldots,\bm{X}((2n_h+1)\Delta t),t,\bm{\theta}_i)-\mu_{\bm{X}^*}(t)\Big)^2 \ ,
\end{multline}
\noindent where $N_s$ is the number of samples forming the Hamiltonian Markov chain that approximates the posterior distribution: $\bm{\theta}_i \sim p(\bm{\theta}|\mathcal{D}) , i=1,\ldots,N_s$. Higher-order moments can also be estimated similarly.

The method is implemented using the JAX library in python (\url{https://github.com/google/jax}) \cite{bradbury2018}. The ensemble forecasts for the surrogate model parameters samples forming the Hamiltonian Markov chain can be conducted in parallel leveraging the \texttt{vmap} and \texttt{pmap} primitives in JAX \cite{bradbury2018}. Simply using these commands allows the parallel computation of designated parts of the code with respect to a batch of different realizations for given variables, instead of rewriting the whole code script in order to parallelize its execution. This option allows leveraging multi-threading on a single processing unit, as well as parallelism across multiple processing units, ultimately enabling computing the forecasts in parallel for the different model parameter samples with effectively no additional computational cost compared to computing forecasts for a single sample of model parameters. This is a major advantage, so that Bayesian trajectories' computation cost is nearly similar to a single deterministic trajectory estimation.

Similarly to the deterministic inference approach, the detailed Bayesian framework can also be applied for standard instantaneous parameterizations described in section \ref{subsec:std_param} by discretizing the parameterized dynamical system (\ref{eq:L96_param_other}) with a time step $\Delta t$.

\section{Results}
\label{sec:res}

The deterministic and Bayesian parameterizations detailed in sections \ref{sec:det_param} and \ref{sec:bayes_param} respectively are applied to the Lorenz '96 model (\ref{eq:L96_X})-(\ref{eq:L96_Y}) in the chaotic regime with an estimated positive maximum Lyapunov exponent for the resolved slow-varying variables. Fully connected neural networks are used as machine learning surrogate models for the parameterizations. All code and data presented in this section are made publicly available at \url{https://github.com/bhouri0412/Hist_Bayesian_Closure}.

\subsection{Problem Setup}
\label{sec:pb_setup}

Similarly to other studies \cite{Gagne2020}, we choose as the number of slow-varying variables $K=8$ and as the number of fast-varying variables per low-frequency variable $J=32$, which gives a Lorenz '96 system of total dimension equal to $264$. The coupling constant is set to $h=1$, the spatial scale ratio to $b=10$, and the temporal scale ration to $c=10$ as chosen in various previous works on Lorenz '96 model \cite{Lorenz96,Wilks2005,Arnold2013,Gagne2020}. The forcing term $F=15$ is taken large enough to ensure chaotic behavior for the resolved variables, with a maximum Lyapunov exponent estimated to be $\lambda_{\rm max}\approx14.80$ for the slow-varying variables. Such a parameter setting results in a model time unit (MTU) that is approximately equivalent to five atmospheric (ATM) days \cite{Lorenz96,Arnold2013,Gagne2020}. 

The history-based parameterization, detailed in sections \ref{sec:det_param} and \ref{sec:bayes_param}, is built using only data of the $K=8$ slow-varying variables for the closure and resulting in a dynamical system that only depends on these slow-varying variables. This parameterization results in an $8$-dimensional dynamical system whose time integration will be compared against the resolved variables states obtained with the ``true'' model of dimension $264$.

The ``true'' model (\ref{eq:L96_X})-(\ref{eq:L96_Y}) is integrated using the RK4 scheme with a time-step equal to $dt=0.005$ from $t=0$ to $t=100 \ \mathrm{MTU} \ (500 \ \mathrm{ATM} \ \mathrm{days})$. In order to account for potential data sparsity in time, for instance such as when using satellite data, we assume that we only have access to the data at each time step equal to $\Delta t=2dt=0.01$. This means that the history-based parameterized model (\ref{eq:L96_param_gen}) is integrated with a time-step equal to $2 \Delta t=0.02$. The resulting training dataset has a total of $9995$ points. We will show that even though the parameterized model's time-step is $4$ times the time-step used to integrate the ``true'' model, the proposed parameterization is capable of returning time forecasts that are faithful and sufficiently accurate compared to the ``true'' model's trajectories in online testing. 

Finally, to model the noise that may be corrupting the observations, each data-point $X_k(t_i) , k=1,\ldots,K, i=0,\ldots,n$ of the training data is altered with an additive Gaussian noise $\mathcal{N}(0,0.03^2\sigma_k^2)$, where $\sigma_k$ is the standard deviation of the training data for the variable $X_k, k=1,\ldots,K$.


\subsection{Model Initialization and Hyperparameter Tuning}

In order to mitigate potential issues of poor initial convergence to local minima in the HMC Markov chain sampler, the inferred neural networks parameters $\bm{\theta}$ are first obtained from the deterministic training detailed in section \ref{sec:det_param}, and then used as a reasonable initial guess of $\bm{\theta}$ for the HMC algorithm. This also promotes robustness and stability in the HMC training process since randomly sampling an initialization for the parameterization may result in exploring a region for the posterior distribution of the model parameter $\bm{\theta}$ for which the time integration of the system may be numerically unstable. The minimization of the deterministic loss (\ref{eq:loss_batch}) is carried out using the Adam optimization \cite{kingma2014}. The deterministic parameterization is exactly equivalent to the MAP estimate for all parameters $\bm{\Theta}$, but excluding the noise precision parameters $\gamma$ and $\lambda$.

The proposed history-based parameterization is evaluated against a standard instantaneous parameterization of the form (\ref{eq:L96_param_other}) where the coupling terms are taken as functions of the slow-varying variables $X_k$, $k=1,\ldots,K$ evaluated only at the current time. The deterministic and Bayesian inference of this non-history-based parameterization are conducted in a similar manner to the framework defined for the proposed history-based one. The only difference in the problem setting regarding these two approaches consists in using the time-step $\Delta t=0.01$ for the non-history-based parameterized model, while the time-step used for the history-based parameterization is $2\Delta t$, as the latter is a constraint imposed by the numerical integration scheme (see Appendix \ref{app_sec:hist_param}). 

History- and non-history-based parameterizations are modeled as fully connected neural networks ($P(\cdot)$ terms of the parameterized DDE (\ref{eq:L96_param_gen}) and $P(\cdot)$ terms of the parameterized ODE (\ref{eq:L96_param_other}). Since the inference problem involves predicting only the next time-step's system state and not a whole time-series, using more complex neural network architectures such as Long Short-Term Memory (LSTM) networks \cite{Hochreiter1997} would not significantly improve the model's prediction accuracy. Besides, for the examples considered in this study, we did not face an issue of vanishing gradients, which may justify using LSTM for instance if encountered in other problems settings. For both parameterizations, the deterministic estimate of the model parameter $\bm{\theta}$ is obtained using $15\times 10^3$ stochastic gradient descent iterations with $n_f=1$ (see Eq. (\ref{eq:loss_batch})), then starting from the inferred parameter $\bm{\theta}$, $30\times 10^3$ stochastic gradient descent iterations are conducted with $n_f>1$ as this allows improving the model accuracy for long-time integration in the online setting. 

Note that the loss for the non-history-based parameterization is slightly different than the history-based one's (\ref{eq:loss_batch}) as the time integration is conducted with a time step of $\Delta t$ instead of $2 \Delta t$. The non-history-based parameterization loss is formulated as follows:
\begin{equation}
    \label{eq:loss_batch_nonhist}\mathcal{J}_{n-h}(\bm{\theta};n_f) = \frac{1}{n_b}\sum_{j\in B} \mathcal{L}\Big(\bm{X}(t_{j+n_f}),\bm{X}^*(t_j+ n_f \Delta t;\bm{\theta})\Big)\ ,
\end{equation}
\noindent where $B$ is a random subset of $\{0,\ldots,n-n_f-1\}$ of size $n_b$. $X^*(t_j+n_f \Delta t;\bm{\theta})$ is the parameterized prediction obtained for the data-point $\bm{X}(t_{j+n_f})$ using the non-history-based parameterized model and starting from the data-point $\bm{X}(t_j)$ , $j=0,\ldots,n-n_f-1$. We conducted a (non-exhaustive) hyperparameters' search for the deterministic training and the corresponding results are detailed in table \ref{tab:hyper_det}.

\begin{table}
\centering
\bgroup
\def\arraystretch{1.5}
\begin{tabular}{|p{0.30\textwidth}|P{0.30\textwidth}|P{0.14\textwidth}|P{0.14\textwidth}|}
\hline
Hyperparameter & Range & Best for history-based parameterization & Best for non-history-based parameterization\\
\hline
Architecture (number of layers x number of nodes per layer) & $\{2\times16,2\times32,4\times64,6\times128,12\times128,8\times256,6\times512\}$  & $6\times128$ & $6\times128$ \\
\hline
Learning rate & $\{10^{-6},10^{-5},10^{-4},10^{-3}\}$ & $10^{-4}$ & $10^{-4}$ \\
\hline
Batch size & $\{128,256,512,1024,2048,4096\}$  & $512$ & $512$ \\
\hline
$n_f$ & $\{2,3,\ldots,9,10\}$ & $5$ & $4$ \\
\hline
$n_h$ & $\{1,2,3,4,5,10\}$ & $2$ & N/A \\
\hline
\end{tabular}
\egroup
\caption{{\em Hyperparameters search for deterministic training of history-based and non-history-based parameterizations.}}
\label{tab:hyper_det}
\end{table}


For the HMC Markov chain sampler, the parameters $\alpha_i$'s and $\beta_i$'s of the prior Gamma distributions for the noise precision parameters $\gamma$ and $\lambda$ (\ref{eq:prior_gamma})-(\ref{eq:prior_lambda}) are taken equal to 1. The deterministic learning precision is used as an initial guess for $\gamma$. This empirical initialization strategy allows accelerating the convergence of the HMC sampler, and hence reduces the overall computational cost to infer the Bayesian parameterization. The HMC step size is taken as equal to $\epsilon_{\mathrm{HMC}}=10^{-4}$, and the number of leapfrog steps to integrate the Hamiltonian dynamics (\ref{eq:HMC_ODE}) is fixed to $L=10$. Higher values for $\epsilon_{\mathrm{HMC}}$ can be considered as long as not too many samples are rejected during model training. More sophisticated HMC samplers with adaptive step-size such as the No U-Turn Sampler could be employed \cite{hoffman2014}, such that the step-size is automatically chosen instead of using a hand-tuned number which may improve the sampler convergence behavior and stabilize the process if needed but we leave this for future work. The HMC Markov Chain is simulated for $4000$ steps. These HMC hyperparameters are kept the same, for a fair comparison, for both history-based and non-history-based parameterizations.

\subsection{Forecasts Results}


The accuracy of the parameterized models forecasts will be measured by evaluating the model root mean square errors (RMSE). The lower the RMSE, the more accurate the forecast. The RMSE for a trajectory defined at the time instances $t_i, i=1,\ldots,N$ is given by:
\begin{equation}
    \label{eq:RMSE} \left.\mathrm{RMSE}(t_N) = \sqrt{ \sum_{i=1}^N ||\bm{X}(t_i)-\bm{X}^*(t_i)||^2_2 } \middle/ \sqrt{ \sum_{i=1}^N ||\bm{X}(t_i)||^2_2 }\right. \ ,
\end{equation}
\noindent where $\bm{X}(t_i),i=1,\ldots,N$ are the ``true'' model points and $\bm{X}^*(t_i),i=1,\ldots,N$ are the parameterized model points obtained by integrating the parameterized model starting from the initial condition (i.e. online forecasting task). The same RMSE can be computed for the closure only (coupling term) by considering the corresponding exact value and the estimates from the parameterized models. Note that the coupling term is learned implicitly in both history- and non-history-based parameterization only from the data on the slow-varying variables.

For the Bayesian parameterizations, the RMSE is computed for the mean and MAP estimates. We also report the fraction of points that are out of the estimated posterior distribution. A data-point is considered out of the estimated posterior distribution if the actual value $X_k(t_i)$ is not in the range $[\mu_{\bm{X}_k^*}(t_i)-2\sigma_{\bm{X}_k^*}(t_i),\mu_{\bm{X}_k^*}(t_i)+2\sigma_{\bm{X}_k^*}(t_i)], k=1,\ldots,K, i=1,\ldots,N$, where $\mu_{\bm{X}_k^*}(t_i)$ and $\sigma_{\bm{X}_k^*}(t_i)$ are the HMC estimates for the first- and second-order statistics of the predicted state $\bm{X}^*_k$ as defined in (\ref{eq:mu_X}) and (\ref{eq:sigma_X}). A lower value for this fraction means a more accurate Bayesian model.

\subsubsection{Deterministic Parameterizations}
\label{subsubsec:res_det}

We first compare the deterministic parameterizations performance in an online setting (i.e. in which the parameterization is coupled to the slow variables equations and integrated forward in time) by solving the parameterized differential equations with the inferred neural networks parameters. The parameterized models are integrated in time till $t=10 \ \mathrm{MTU} = 50 \ \mathrm{ATM \ days}$ starting from the first and last point of the training dataset. 

\begin{figure}
     \centering
     \begin{subfigure}[b]{0.45\textwidth}
         \centering
         \includegraphics[width=\textwidth]{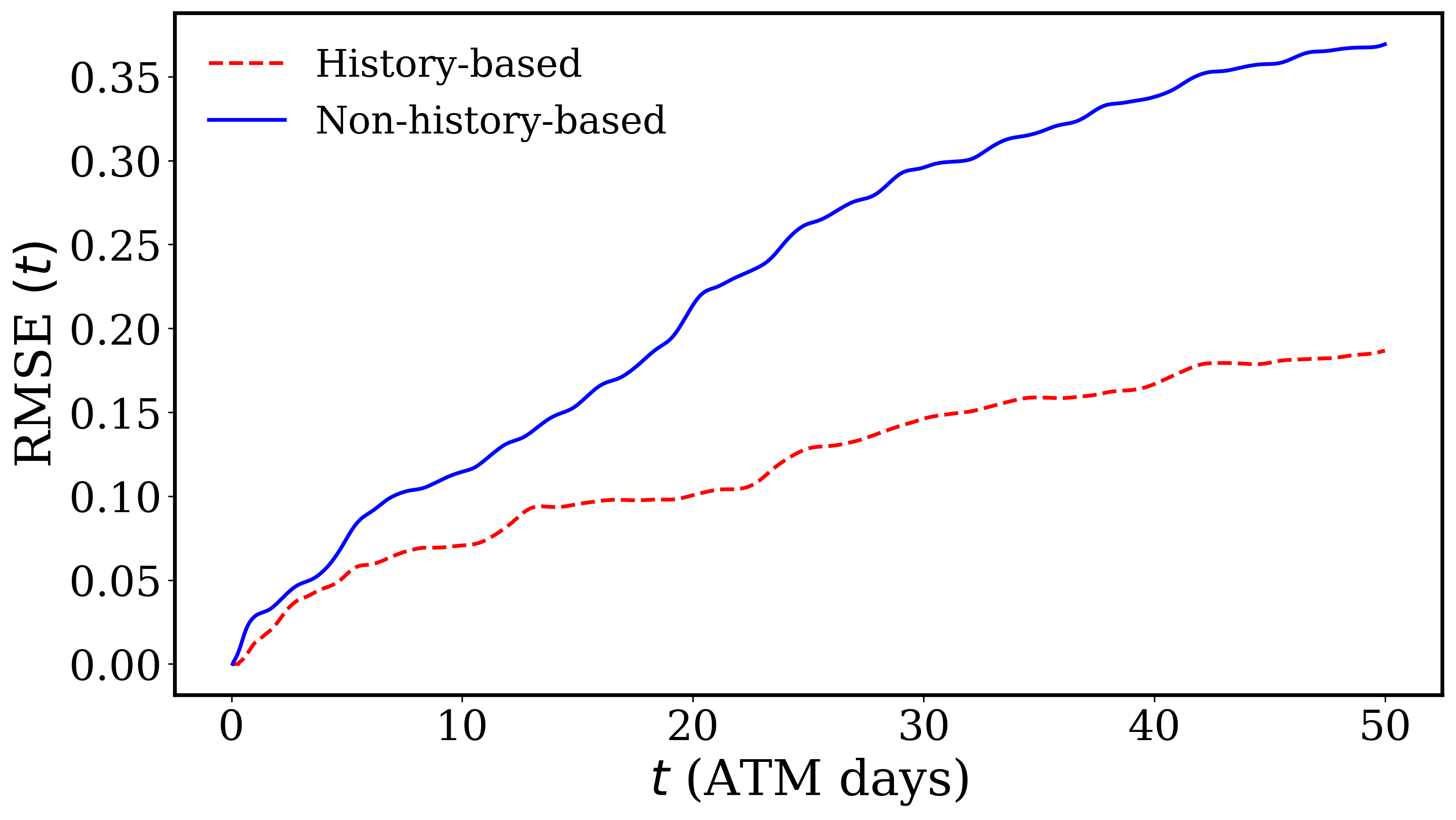}
         \caption{{\em Temporal evolution of the RMSE starting from the first training point.}}
         \label{fig:det_RMSE_X_first}
     \end{subfigure}
     \begin{subfigure}[b]{0.45\textwidth}
         \centering
         \includegraphics[width=\textwidth]{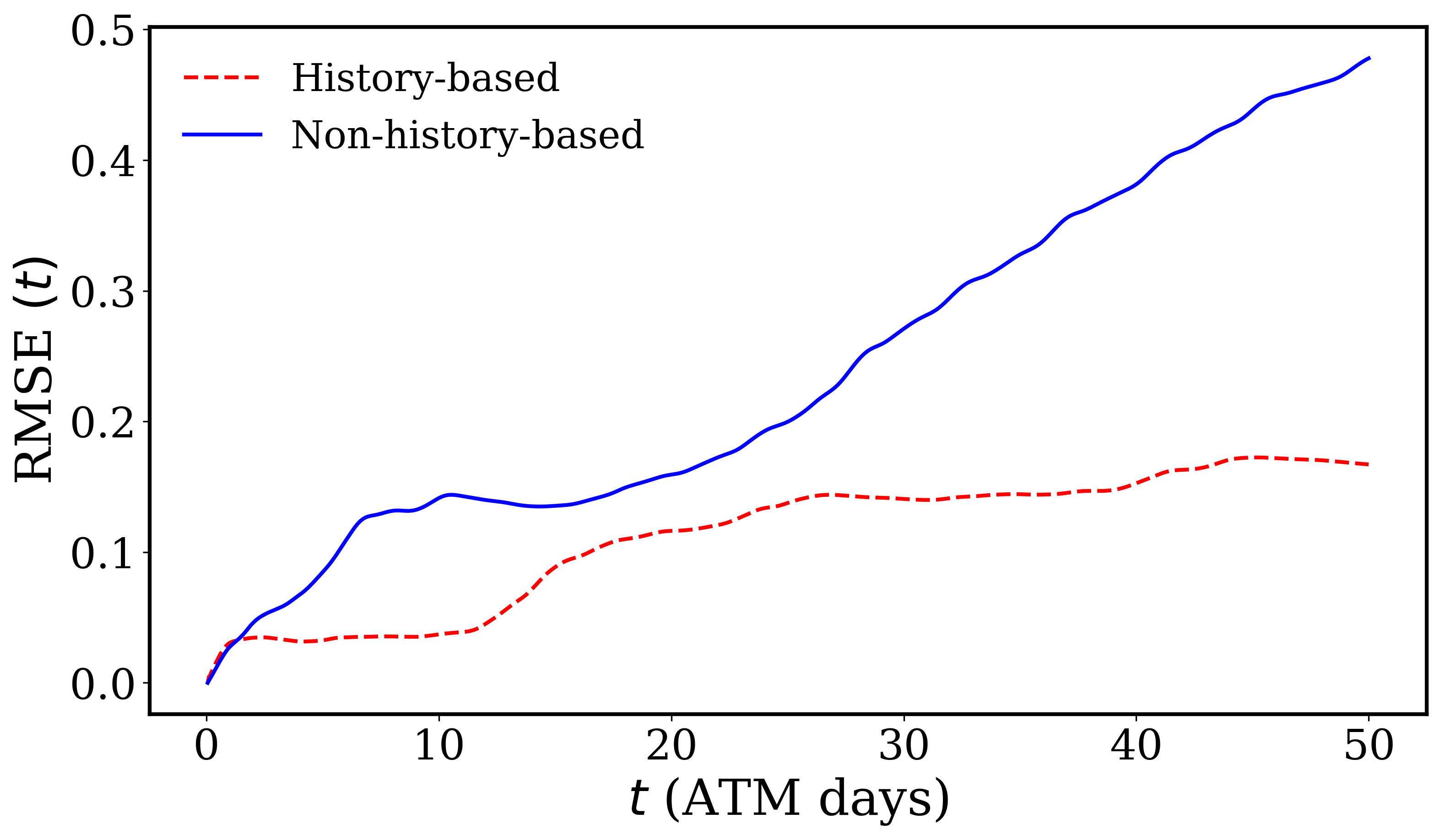}
        \caption{{\em Temporal evolution of the RMSE starting from the last training point.}}
         \label{fig:det_RMSE_X_last}
     \end{subfigure}
        \caption{{\em Temporal evolution of the deterministic parameterizations' RMSEs for the slow-varying variables.}}
        \label{fig:det_RMSE_X}
\end{figure}

Table \ref{tab:det_RMSE} gathers the final RMSE for the slow-varying variables and the closure term for both parameterizations. Figure \ref{fig:det_RMSE_X} shows the temporal evolution of the slow-varying variables' RMSE for both parameterizations starting form the first and last training points. These results clearly prove the significant improvement obtained by using the history-based parameterization compared to a classical approach where the instantaneous parameterization only depends on the state of the resolved variables at the current time. In particular, the RMSE is always lower for the history-based parameterization at any extrapolation time, and even when errors accumulate after time integration is carried out for longer intervals, the accumulation is significantly lower for the history-based parameterization than the instantaneous one as shown in figure \ref{fig:det_RMSE_X}.

\begin{table}
\centering
\bgroup
\def\arraystretch{1.5}
\begin{tabular}{|p{0.17\textwidth}|P{0.17\textwidth}|P{0.17\textwidth}|P{0.17\textwidth}|P{0.17\textwidth}|}
\hline
Parameterization & RMSE for $\bm{X}$ starting from the first training point & RMSE for $\bm{X}$ starting from the last training point & RMSE for closure starting from the first training point parameterization & RMSE for closure starting from the last training point parameterization\\
\hline
Non-history-based &$0.370$ & $0.478$ & $0.112$ & $0.112$ \\
\hline
History-based & $0.187 $  & $0.167 $ & $0.107$ & $0.106$ \\
\hline
\end{tabular}
\egroup
\caption{{\em Deterministic parameterizations performance in an online task:} final RMSEs for history- and non-history-based parameterizations}
\label{tab:det_RMSE}
\end{table}

\begin{figure}
\centering
\includegraphics[width=0.8\textwidth]{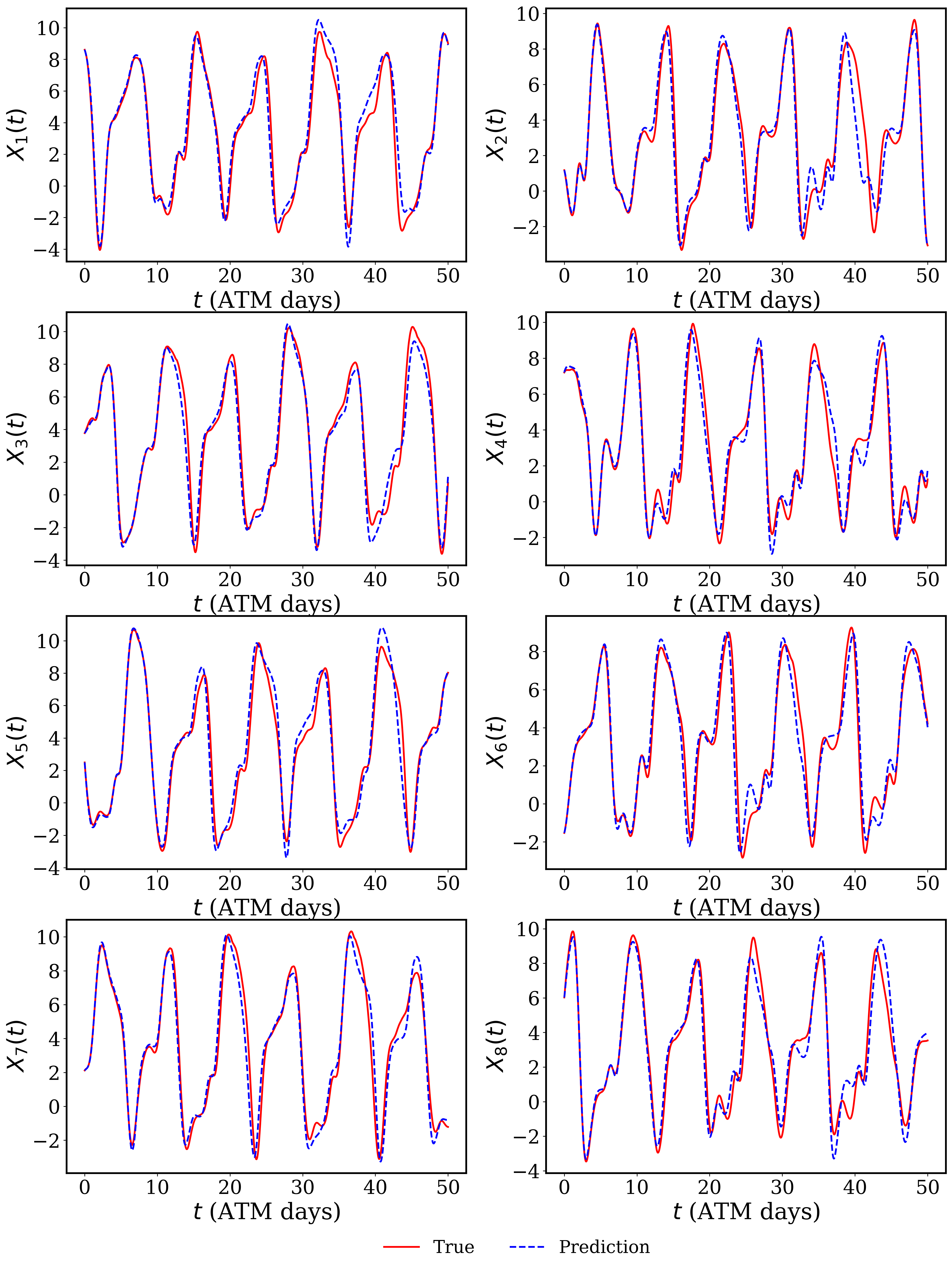}
\caption{{\em Online predictions for the resolved slow-varying variables using the deterministic history-based parameterization starting from the last training point:} Predictions correspond to solutions of the parameterized DDE (\ref{eq:L96_param_gen}), while true trajectories correspond to solutions of the ``true'' model (\ref{eq:L96_X})-(\ref{eq:L96_Y}).}
\label{fig:hist_X_det}
\end{figure}

\begin{figure}
\centering
\includegraphics[width=0.8\textwidth]{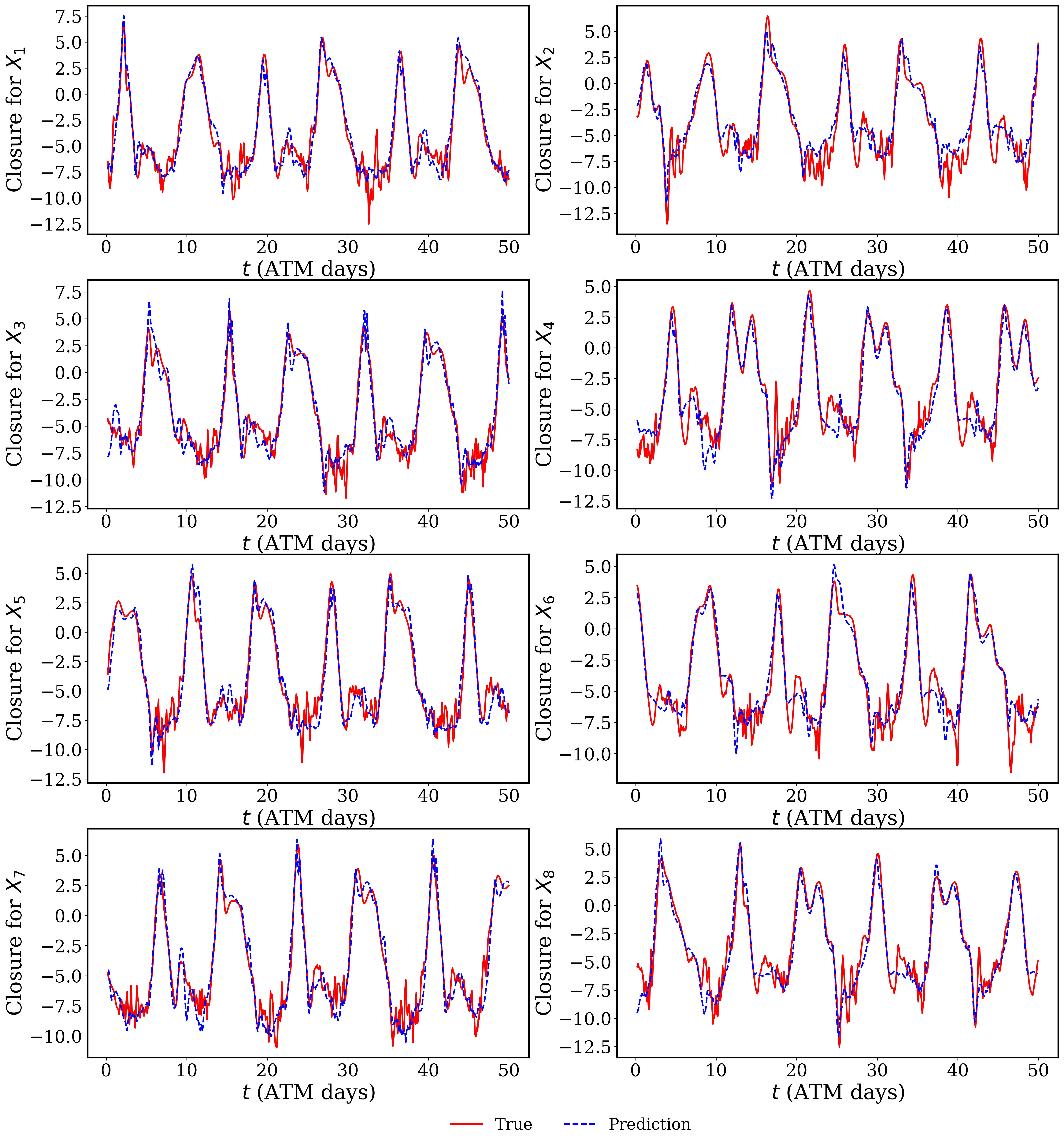}
\caption{{\em Online predictions for the closure terms using the deterministic history-based parameterization starting from the last training point:} Predictions correspond to the $P(\cdot)$ terms of the parameterized DDE (\ref{eq:L96_param_gen}), while true trajectories correspond to the coupling term in (\ref{eq:L96_X}).}
\label{fig:hist_closure_det}
\end{figure}

Figures \ref{fig:hist_X_det} and \ref{fig:hist_closure_det} show the online predictions for the slow-varying variables and the closure terms respectively, using the deterministic history-based parameterization and starting from the last training point. The good agreement between the true trajectory and the parameterized model prediction confirms the ability of the proposed framework to learn a parameterization that is stable and also sufficiently accurate when tested online. It also shows that the model is capable of implicitly inferring the closure term correctly, without using any data or information on closure or on the unresolved fast-varying variables, but by only being trained on the sparse and noisy trajectories of the resolved variables. For reference, we provide online predictions for the slow-varying variables using the deterministic non-history-based parameterization and starting from the last training point in Appendix \ref{app:nonhist}, which shows the prediction mismatch compared to the true trajectory due to the use of a parameterization depending only on the current state of the slow-varying variables.

An interesting observation regarding the learned parameterization is that the actual closure displays higher frequencies than the inferred trajectories from the parameterization as shown in figure \ref{fig:hist_closure_det}. This is due to the fact that the actual closure terms have a higher frequency than the resolved slow-varying variables since they depend on the fast-varying variables. On the other hand, the inferred closure terms show a lower frequency that is closer to the frequency of the slow-varying variables than to the fast-varying variables' one. This result is coherent with the fact that the model is trained to match and predict the slow-varying variables, while no data or information on the closure terms or the fast-varying (high-frequency) variables is available. Although some high frequencies of the closure terms are filtered out by the parameterization given the problem setup considered, the parameterized model is still able to provide stable and accurate online temporal predictions for the resolved variables.

\subsubsection{Bayesian Parameterizations}
\label{subsubsec:res_Bayes}

We now compare the Bayesian parameterizations performance in an online setting by solving the parameterized ODE (\ref{eq:L96_param_other}) and DDE (\ref{eq:L96_param_gen}) using the inferred parameters for the neural networks. The parameterized models are integrated in time till $t=10 \ \mathrm{MTU} = 50 \ \mathrm{ATM \ days}$ starting from the first point of the training dataset, and then from the last point of the training dataset. 

\begin{figure}
     \centering
     \begin{subfigure}[b]{0.45\textwidth}
         \centering
         \includegraphics[width=\textwidth]{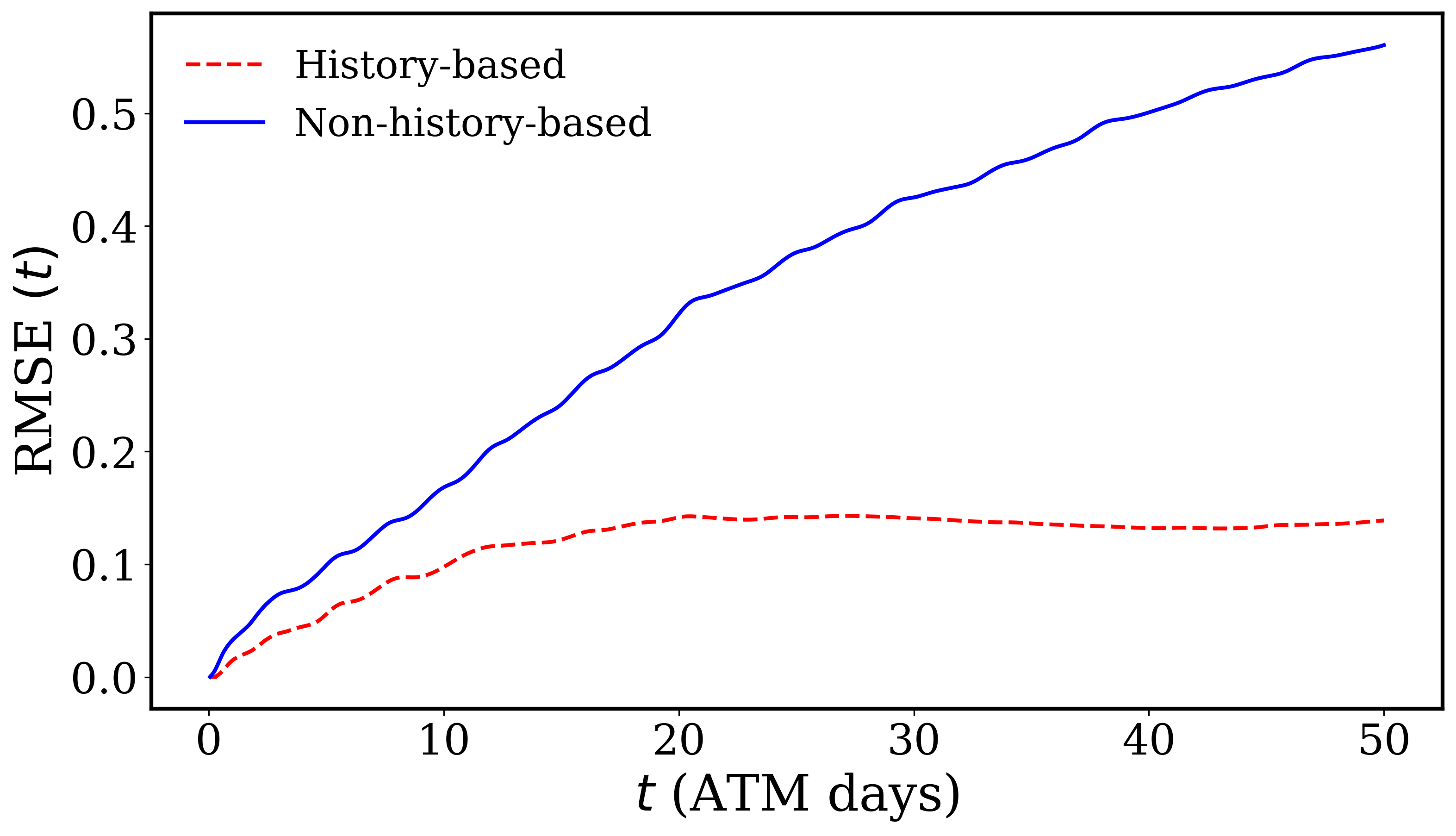}
         \caption{{\em Temporal evolution of the RMSE starting from the first training point.}}
         \label{fig:bayes_RMSE_mean_X_first}
     \end{subfigure}
     \begin{subfigure}[b]{0.45\textwidth}
         \centering
         \includegraphics[width=\textwidth]{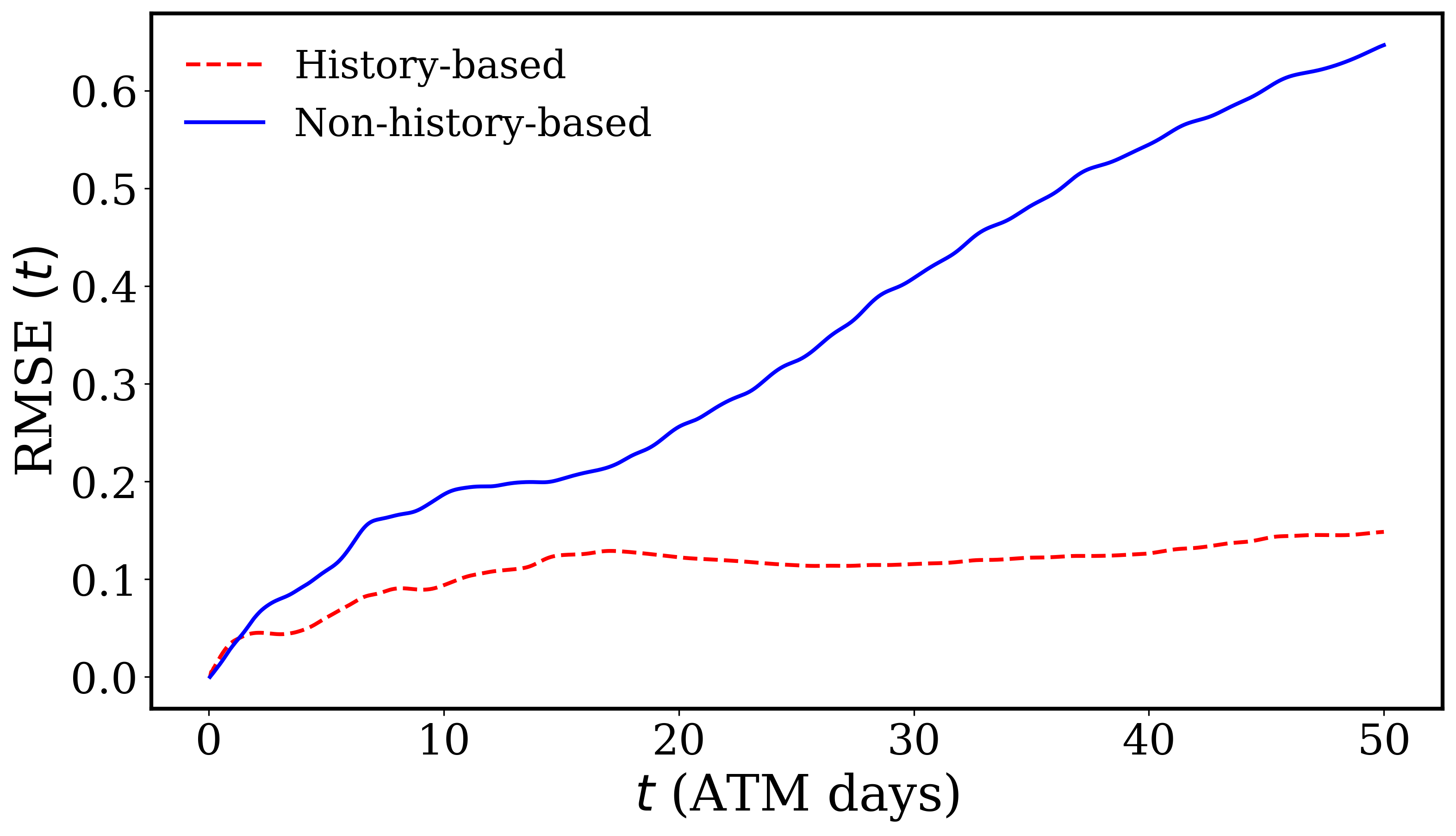}
        \caption{{\em Temporal evolution of the RMSE starting from the last training point.}}
         \label{fig:bayes_RMSE_mean_X_last}
     \end{subfigure}
        \caption{{\em Temporal evolution of the Bayesian parameterizations' RMSEs using mean estimates for the slow-varying variables.}}
        \label{fig:bayes_RMSE_mean_X}
\end{figure}

\begin{figure}
     \centering
     \begin{subfigure}[b]{0.45\textwidth}
         \centering
         \includegraphics[width=\textwidth]{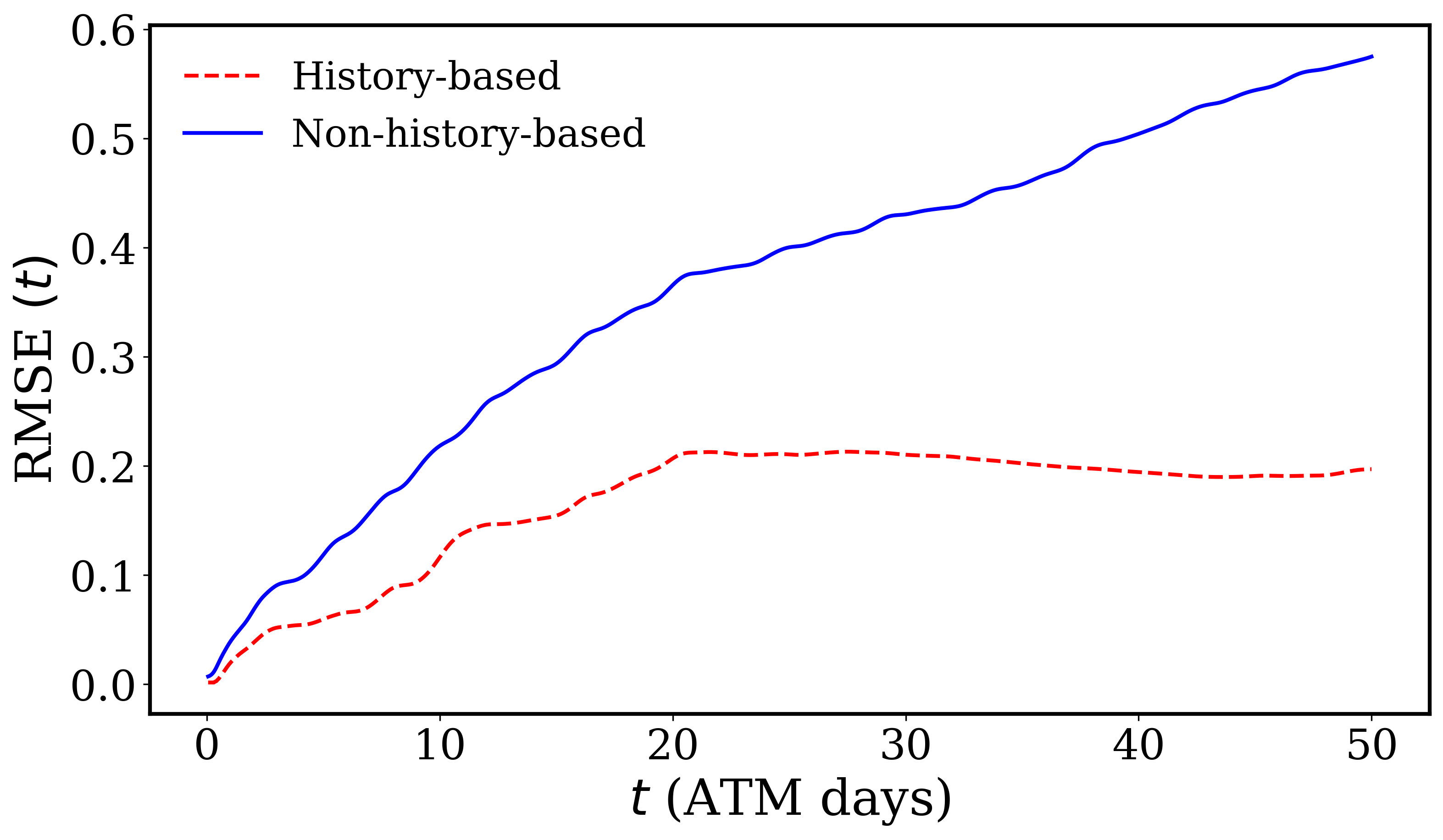}
         \caption{{\em Temporal evolution of the RMSE starting from the first training point.}}
         \label{fig:bayes_RMSE_MAP_X_first}
     \end{subfigure}
     \begin{subfigure}[b]{0.45\textwidth}
         \centering
         \includegraphics[width=\textwidth]{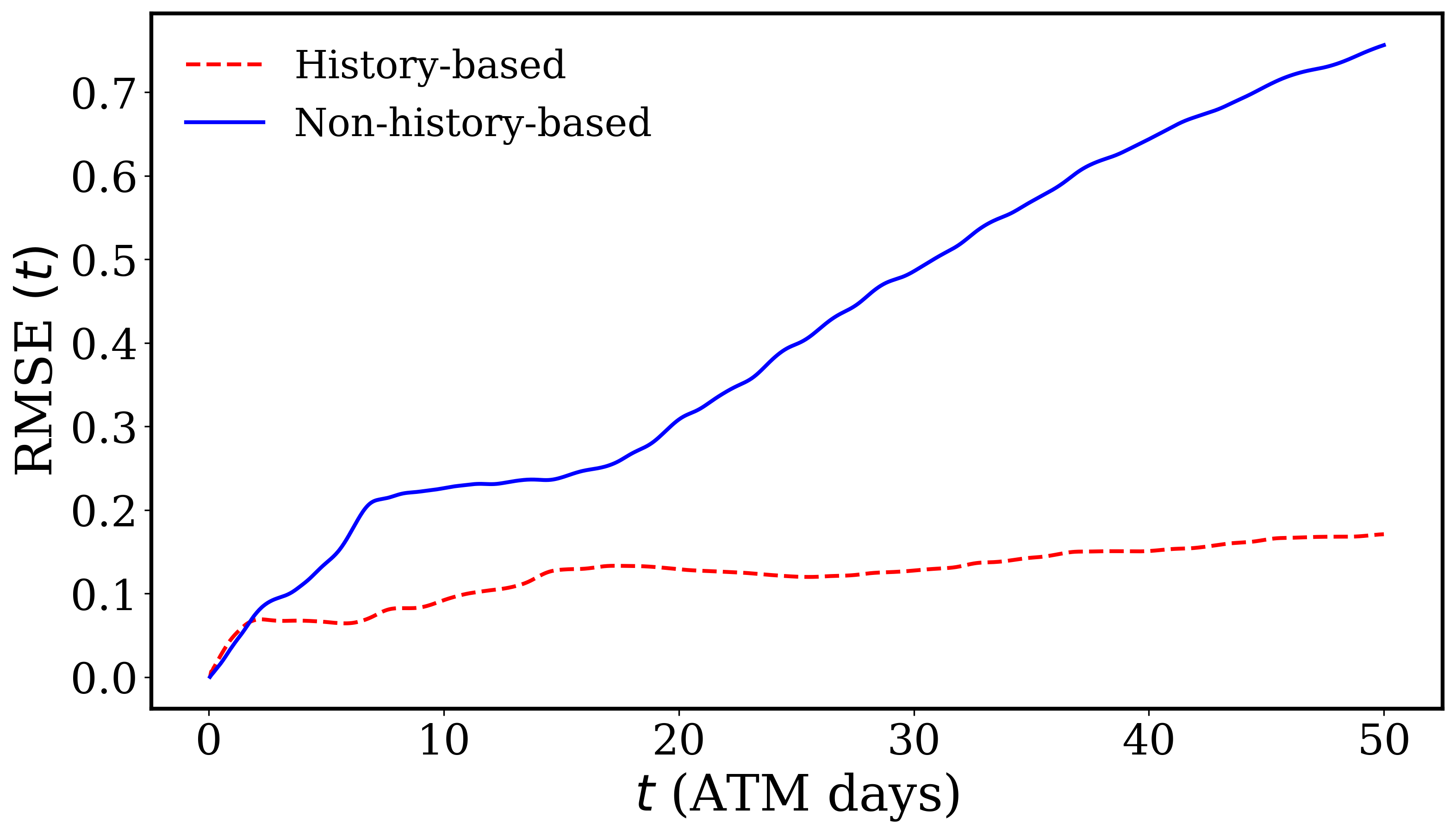}
        \caption{{\em Temporal evolution of the RMSE starting from the last training point.}}
         \label{fig:bayes_RMSE_MAP_X_last}
     \end{subfigure}
        \caption{{\em Temporal evolution of the Bayesian parameterizations' RMSEs using MAP estimates for the slow-varying variables.}}
        \label{fig:bayes_RMSE_MAP_X}
\end{figure}

Tables \ref{tab:Bayes_RMSE_mean} and \ref{tab:Bayes_RMSE_MAP} present the final RMSE for the slow-varying variables and the closure term for both parameterizations using the mean and MAP estimates respectively. Table \ref{tab:Bayes_out_of_dist} presents the fraction of points out of the estimated posterior distributions for the slow-varying variables and the closure term for both parameterizations. Figures \ref{fig:bayes_RMSE_mean_X} and \ref{fig:bayes_RMSE_MAP_X} show the temporal evolution of the slow-varying variables' RMSE for both Bayesian parameterizations starting form the first and last training points using mean and MAP estimates respectively. 

As observed for the deterministic parameterizations, these results prove the significant improvement obtained by using the history-based parameterization compared to a classical approach. For the non-history-based approach, the error introduced by the deterministic parameterization could not be solved by the HMC sampler. Whether the approach is deterministic or Bayesian, requiring the parameterized differential equation to provide trajectories that are sufficiently accurate and close to the ones of the original ``true'' equation does not seem to be possible if the parameterization only depends on the variable states at the current time, but this is substantially improved when relying on a history-based parameterization. Similarly to the deterministic parameterizations, the RMSEs are always lower for the history-based parameterization at any extrapolation time as shown in figures \ref{fig:bayes_RMSE_mean_X} and \ref{fig:bayes_RMSE_MAP_X}.

\begin{table}
\centering
\bgroup
\def\arraystretch{1.5}
\begin{tabular}{|p{0.17\textwidth}|P{0.17\textwidth}|P{0.17\textwidth}|P{0.17\textwidth}|P{0.17\textwidth}|}
\hline
Parameterization & RMSE for $\bm{X}$ starting from the first training point & RMSE for $\bm{X}$ starting from the last training point & RMSE for closure starting from the first training point parameterization & RMSE for closure starting from the last training point parameterization\\
\hline
Non-history-based & $0.561$  & $0.647$ & $0.133$ & $0.132$ \\
\hline
History-based & $0.139 $  & $0.149 $ & $0.112 $ & $0.111 $ \\
\hline
\end{tabular}
\egroup
\caption{{\em Bayesian parameterizations performance in an online task using the mean estimates:} final RMSEs for history- and non-history-based parameterizations}
\label{tab:Bayes_RMSE_mean}
\end{table}

\begin{table}
\centering
\bgroup
\def\arraystretch{1.5}
\begin{tabular}{|p{0.17\textwidth}|P{0.17\textwidth}|P{0.17\textwidth}|P{0.17\textwidth}|P{0.17\textwidth}|}
\hline
Parameterization & RMSE for $\bm{X}$ starting from the first training point & RMSE for $\bm{X}$ starting from the last training point & RMSE for closure starting from the first training point parameterization & RMSE for closure starting from the last training point parameterization\\
\hline
Non-history-based & $0.575$  & $0.757$ & $0.133$ & $0.132$ \\
\hline
History-based & $0.197 $  & $0.171 $ & $0.127 $ & $0.131 $ \\
\hline
\end{tabular}
\egroup
\caption{{\em Bayesian parameterizations performance in an online task using the MAP estimates:} final RMSEs for history- and non-history-based parameterizations}
\label{tab:Bayes_RMSE_MAP}
\end{table}

\begin{table}
\centering
\bgroup
\def\arraystretch{1.5}
\begin{tabular}{|p{0.17\textwidth}|P{0.17\textwidth}|P{0.17\textwidth}|P{0.17\textwidth}|P{0.17\textwidth}|}
\hline
Parameterization & Fraction of $\bm{X}$ points out of the posterior starting from the first training point & Fraction of $\bm{X}$ points out of the posterior starting from the last training point & Fraction of closure estimate points out of the posterior starting from the first training point & Fraction of closure estimate points out of the posterior starting from the last training point\\
\hline
Non-history-based & $0.741$  & $0.666$ & $0.719$ & $0.724$ \\
\hline
History-based & $0.143 $  & $0.130 $ & $0.468 $ & $0.478 $ \\
\hline
\end{tabular}
\egroup
\caption{{\em Bayesian parameterizations performance in an online task using the fraction of points that are out of the estimated posterior distribution:} comparison between history- and non-history-based parameterizations}
\label{tab:Bayes_out_of_dist}
\end{table}

\begin{figure}
\centering
\includegraphics[width=0.8\textwidth]{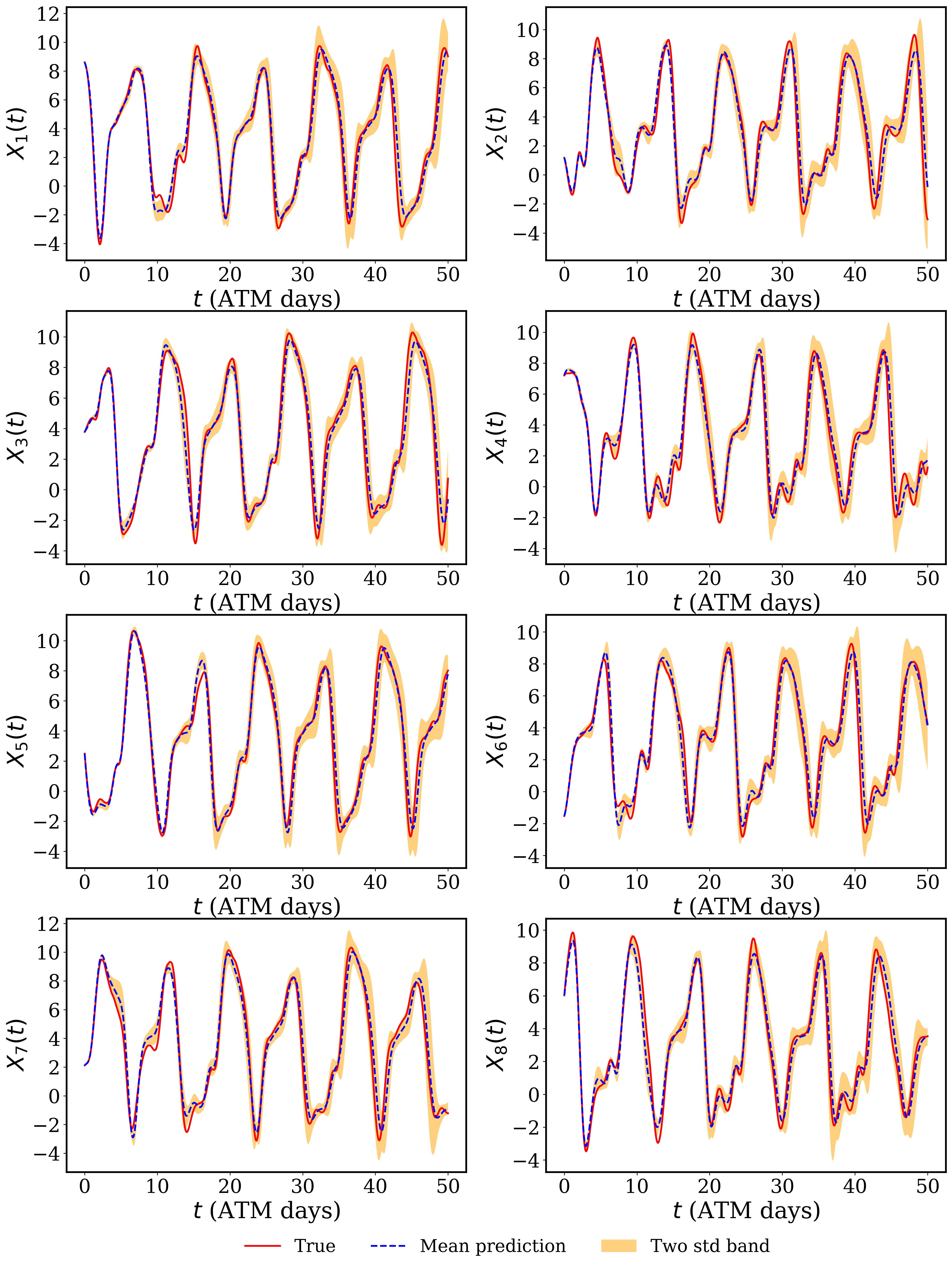}
\caption{{\em Online predictions for the resolved slow-varying variables using the Bayesian history-based parameterization starting from the last training point:} Predictions correspond to the ensemble estimates of the parameterized DDE (\ref{eq:L96_param_gen}) solutions using the HMC Markov Chain parameters, while true trajectories correspond to solutions of the ``true'' model (\ref{eq:L96_X})-(\ref{eq:L96_Y}).}
\label{fig:hist_mean_X}
\end{figure}

\begin{figure}
\centering
\includegraphics[width=0.8\textwidth]{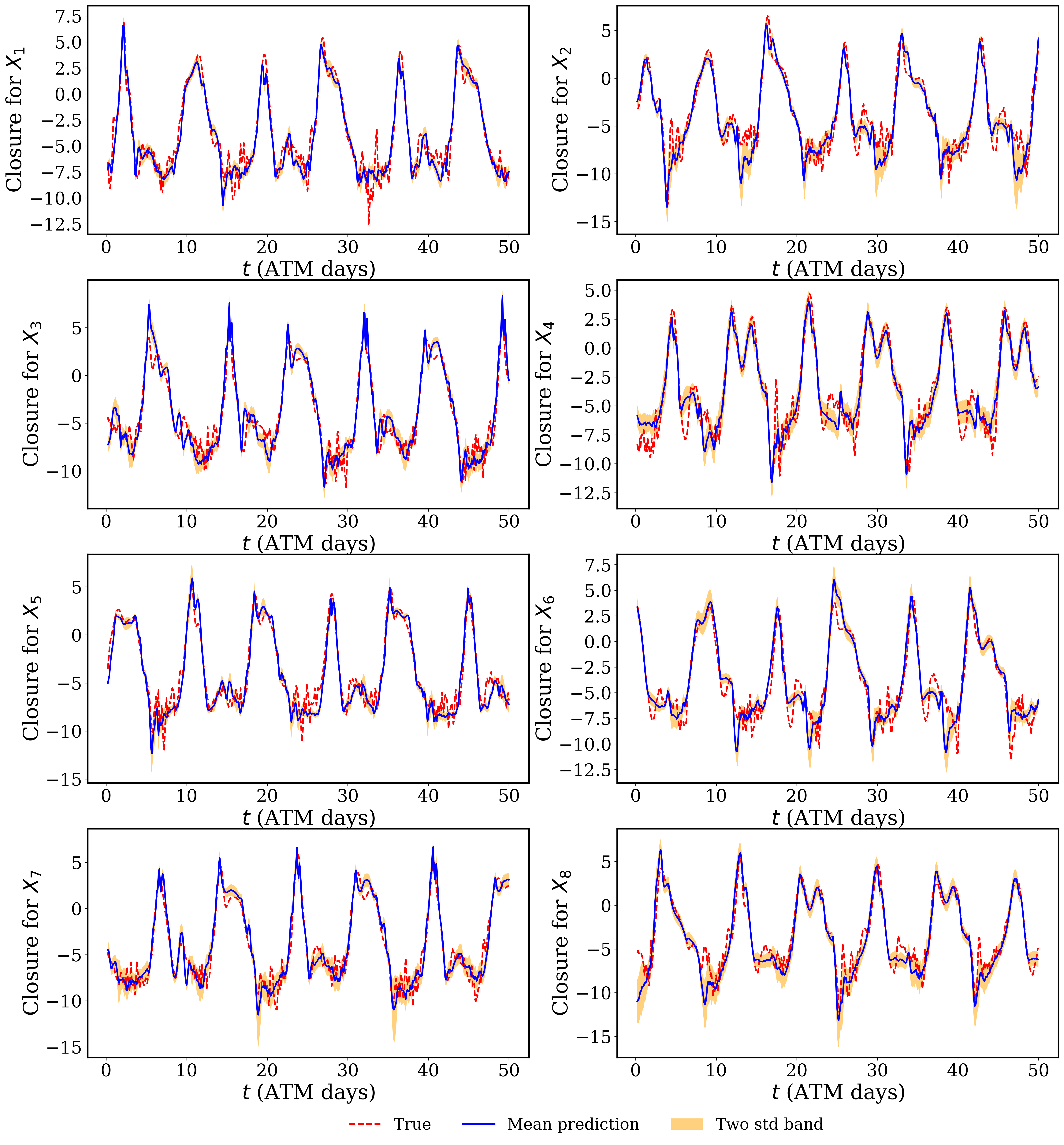}
\caption{{\em Online predictions for the closure terms using the Bayesian history-based parameterization starting from the last training point:} Predictions correspond to the ensemble estimates of the $P(\cdot)$ terms of the parameterized DDE (\ref{eq:L96_param_gen}) using the HMC Markov Chain parameters, while true trajectories correspond to the coupling term in (\ref{eq:L96_X}).}
\label{fig:hist_mean_closure}
\end{figure}

Figures \ref{fig:hist_mean_X} and \ref{fig:hist_mean_closure} show the online predictions for the slow-varying variables and the closure terms respectively, using the Bayesian history-based parameterization and starting from the last training point. The predictions correspond to the ensemble estimates of the parameterized DDE (\ref{eq:L96_param_gen})'s solutions and the $P(\cdot)$ terms using the HMC Markov Chain parameters. For reference, we provide the slow-variables' predictions corresponding to the ensemble estimates using the HMC Markov Chain parameters in Appendix \ref{app:hist}. 

As observed for the deterministic approach, the good agreement between the true trajectory and the Bayesian parameterized model prediction confirms the ability of the proposed framework to learn a parameterization that is stable and also accurate when tested online, without having to rely on the unresolved fast-varying variables. We also provide the online predictions for the slow-varying variables using the Bayesian non-history-based parameterization starting from the last training point in Appendix \ref{app:nonhist}, which shows the prediction mismatch compared to the true trajectory due to the instantaneous parameterization depending only on the current state of the slow-varying variables. 

For the Bayesian history-based parameterization, the uncertainty quantification obtained via sampling from the joint posterior distribution over all model parameters is able to well capture the exact trajectory of the ``true'' model which is solved with a time-step $4$ times smaller than the time-step used for the parametrized system. As expected, the estimated uncertainty is larger as the extrapolation time interval increases and in regions where we have significant variations (e.g. for the variable $X_4$ around $t=30  \ \mathrm{ATM} \ \mathrm{days}$ and the variable $X_8$ around $t = 40 \mathrm{ATM} \ \mathrm{days}$, as shown in figure \ref{fig:hist_mean_X}). These results are consistent with the problem setup, as in these regions sources of error are more significant given the chaotic nature of the dynamical system and the numerical error's accumulation when integrating for longer time periods. Nonetheless, even in these regions, the uncertainty estimates always capture the exact trajectory within its posterior distribution and the mean follows the true trajectory with sufficient accuracy. 

The estimated posterior distribution is also meaningful for the closure terms as we observe a higher uncertainty in regions where we have higher frequency variations, as shown in figure \ref{fig:hist_mean_closure}. As explained for the deterministic parameterization results, the inferred closure term displays lower frequencies compared to the true closure, since the model is trained using only the slow-varying variables. Nonetheless, the Bayesian formalism allows discovering regions of higher uncertainties including those corresponding to higher frequency variations. This inherent capability of quantifying the uncertainty of the model's predictions is a key advantage of the Bayesian formalism.  Given the results presented of the proposed Bayesian parameterization, the HMC-based framework can identify and quantify these different sources of error and returns consistent uncertainty quantification. Quantification of individual sources of uncertainty is studied in more depth in section \ref{sec:uncert_quan}.

An interesting question that can be tackled for any parameterization is: ``what is the parameterization actually learning?''. Although the parameterization is designed to learn some specific terms (e.g. the coupling term of the Lorenz '96 in this work), there is no constraint to guarantee that the learning process is limited to these quantities.

In the problem setup considered in this work, data sparsity is a major source of error that seems to be accounted for and resolved by the history-based parameterization as shown in the results presented in sections \ref{subsubsec:res_det} and \ref{subsubsec:res_Bayes}. Indeed, the ``true'' model (\ref{eq:L96_X})-(\ref{eq:L96_Y}) is integrated using RK4 with a time-step equal to $dt=0.005$, while the history-based parameterized model is trained on data with a time-step equal to $\Delta t=2dt=0.01$. This means that the history-based parameterized model is integrated with a time-step equal to $2 \Delta t=0.02$, which is $4$ times the time-step used to integrate the ``true'' model. 

A simple exercise to verify the impact of the parameterization on correcting the numerical error introduced by the finer temporal resolution considered for the parameterized model is to solve the ``true'' model with a time-step equal to $2 \Delta t=0.02$ and compare its RMSE with the error obtained for the parameterized model, while considering the solution of the ``true'' model with a time-step equal to $dt=0.005$ as reference.

\begin{figure}
     \centering
     \includegraphics[width=.7\textwidth]{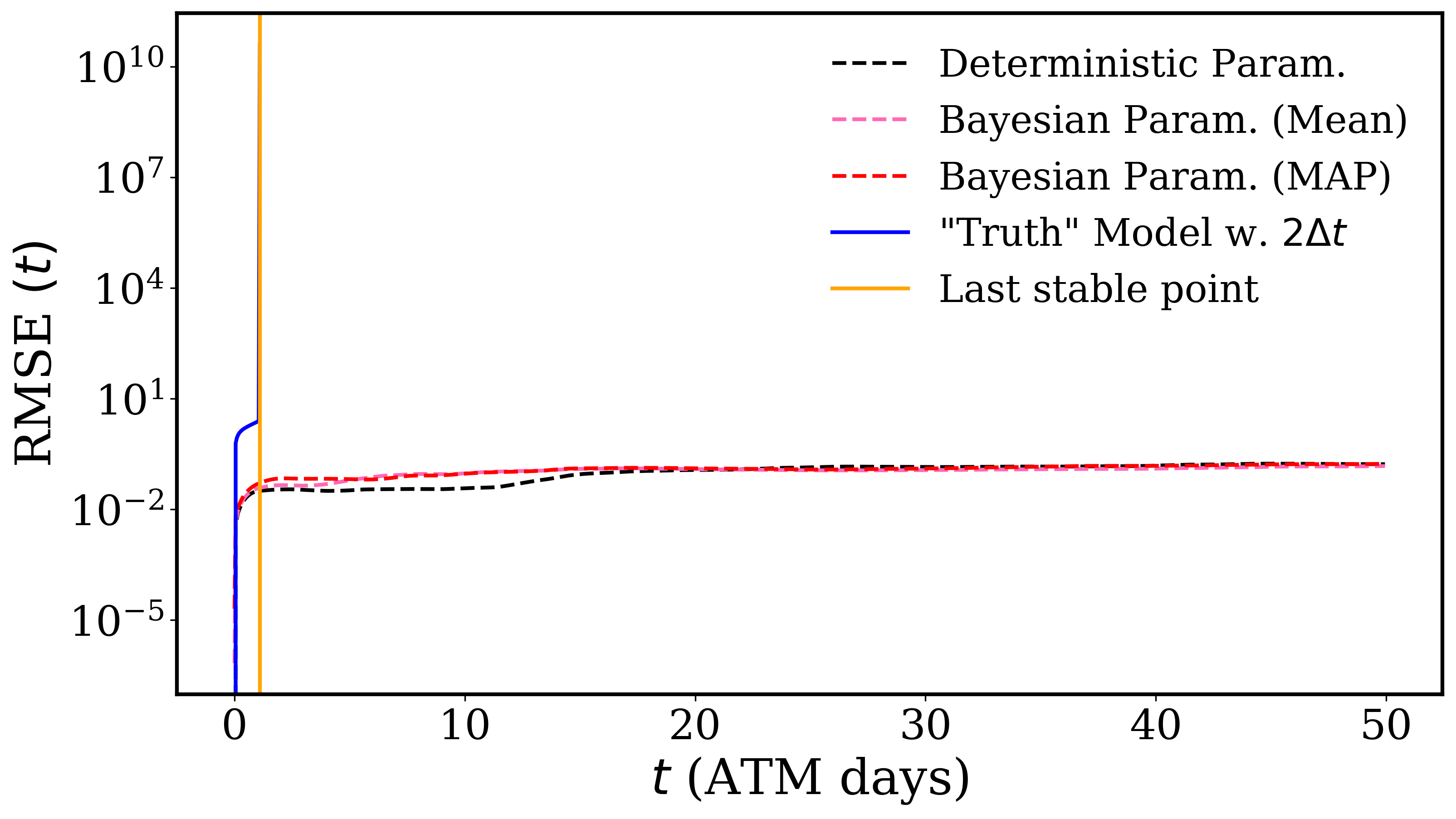}
    \caption{{\em Temporal evolution of the RMSEs starting from the first training point for the history-based-parameterized model (\ref{eq:L96_param_gen}) and the ``true'' model (\ref{eq:L96_X})-(\ref{eq:L96_Y}), both being solved with a time-step equal to $2 \Delta t=0.02$.}}
    \label{fig:RMSE_truth_int}
\end{figure}

Solving the  ``true'' model with a time-step equal to $2 \Delta t=0.02$
returns a solution that diverges at $t = 1.1 \mathrm{ATM} \ \mathrm{days}$. We show the corresponding RMSE in figure \ref{fig:RMSE_truth_int} which also includes the RMSEs of the history-based-parameterized model (\ref{eq:L96_param_gen}) solved with the same time-step equal to $2 \Delta t=0.02$. This result means that the history-based parameterization does not only learn the coupling term, but also improves the computational stability and accuracy of the learned dynamical system. 

\subsection{Uncertainty Quantification}
\label{sec:uncert_quan}

This inherent capability of quantifying the uncertainty of the model's predictions is the core advantage of the Bayesian formalism. For the Lorenz '96 parameterization considered in this work, these sources of uncertainty consist mainly in (1) the chaotic and/or stochastic nature of the dynamical system, (2) the limitation introduced by the parameterization compared to the ``true'' model, (3) the quality of the data given the noise level and temporal sparsity and (4) the discretization of the continuous differential equations. This last source of uncertainty is inherent to the problem considered and cannot be ``synthetically'' controlled (for instance using model intervention) since there is no closed form analytical solution to the ``true'' model (\ref{eq:L96_X})-(\ref{eq:L96_Y}). However, we can synthetically alter the first three sources of uncertainties listed above and quantify the related uncertainty estimates returned by the instantaneous and history-based parameterizations. 

The uncertainty related to the chaotic nature of the Lorenz '96 model can be controlled by varying the forcing term $F$ the ``true'' model (\ref{eq:L96_X})-(\ref{eq:L96_Y}). Larger values for $F$ result in more chaotic behavior of the dynamical system. The uncertainty related to the limitation introduced by the parametrization scheme can be quantified by comparing the uncertainty estimates obtained with the Bayesian history- and non-history-based parameterizations. Finally, the noise magnitude defined by the variance of the Gaussian noise added to the observations is used to modify the uncertainty related to the quality of the available data.

Uncertainty estimates are quantified by computing the mean relative standard deviation defined as 
\begin{equation}
    \label{eq:rel_sigma}\sigma_r=\frac{1}{nK}\sum_{k=1}^K\sum_{i=1}^n \frac{\sigma_{\bm{X}_k^*}(t_i)}{\bm{X}_k(t_i)} ,
\end{equation}
\noindent where $\sigma_{\bm{X}_k^*}$ is the variance defined in (\ref{eq:sigma_X}) and estimated by HMC algorithm and $\bm{X}$ is the resolved variable as solution to the ``true'' model (\ref{eq:L96_X})-(\ref{eq:L96_Y}).

Table \ref{tab:uncert_quan} gathers the uncertainty quantification obtained with the HMC algorithm for different scenarios. For fixed noise level and parameterization scheme, both Bayesian parameterization approaches are capable of returning meaningful uncertainty quantification with respect to the chaotic nature of the dynamical system. Uncertainty estimates systematically increases as the dynamical system considered has a more pronounced chaotic behavior (i.e. higher $F$ values). 

For a given forcing value, the proposed Bayesian history-based parameterization systematically returns meaningful uncertainty quantification with respect to the level of noise altering the observational data. For any forcing value, higher noise magnitudes results in a higher uncertainty estimation for the the forecasts of the resolved variables as shown in table \ref{tab:uq_hist}. It is also interesting to notice that the uncertainty quantification returned by the Bayesian history-based parameterization is coherent with the multiple sources of uncertainty since as the forcing value is higher, the effect of the noise level on the uncertainty is lower. This result suggests that for high values of forcing (around $F=20$), the major source of uncertainty is related to the chaotic nature of the dynamical system rather than to the noise level. However, the Bayesian instantaneous (non-history-based) parameterization fails to skillfully quantify the uncertainty related to the noise corrupting the observational data as detailed in table \ref{tab:uq_non_hist} since for a given forcing value, its uncertainty estimates is lower for the higher noise level. This behavior suggests that the parameterization limitation introduced by the non-history-based scheme is a major source of uncertainty which outweighs the uncertainty introduced by the noise corrupting the observational data. Indeed, the uncertainty estimation obtained with the instantaneous parameterization for the noise level $10\%$ can be explained by a relative uncertainty collapse compared to the estimates obtained for the noise level $3\%$. These results highlight the importance of relying on the history-based parameterization scheme since it does not only contribute to reduie the model's error, but it also allows a more skillful uncertainty quantification.

Finally, by considering the noise level $3\%$ for which the uncertainty estimates for the non-history-based scheme seems reliable, table \ref{tab:uncert_quan} shows that the HMC algorithm applied to the parameterization schemes returns coherent uncertainty estimates with respect to the limitations introduced by the parameterization since the history-based scheme is more expressive than the non-history-based one, and hence is more capable of explaining and fitting the resolved variables trajectories returned by the ``true'' model (\ref{eq:L96_X})-(\ref{eq:L96_Y}). This difference results in a lower uncertainty in the predictions made with the Bayesian history-based parameterization than with the Bayesian non-history-based one for the  same problem setup.

All these results show that, unlike the non-history-based scheme, the Bayesian history-based parameterization has the ability to accurately quantify several sources of uncertainty by inferring appropriate posterior distributions over the surrogate model and precision parameters. Uncertainty quantification for these inferred parameters is then propagated through the parameterization predictions, and finally through the forecasts of the resolved variables as detailed in table \ref{tab:uncert_quan}. 

\begin{table}
\centering
\begin{subtable}{0.55\textwidth}
\centering
\def\arraystretch{1.5}
\begin{tabular}[t]{|p{0.225\textwidth}|P{0.2\textwidth}|P{0.2\textwidth}|}
 \hline %
 \backslashbox{$F$}{Noise} &   $3\%$ & $10\%$ \\
 \hline
    5 &   $1.16\times10^{-2}$ &   $9.42\times10^{-3}$ \\
 \hline                                                 
   15  &   $5.31$ & $8.91\times10^{-1}$  \\
 \hline   
    20 &   $2.78\times10$ & $1.42\times10$  \\
 \hline
\end{tabular}
\caption{{\em Non-history-based parameterization}}
\label{tab:uq_non_hist}
\end{subtable}
\begin{subtable}{0.55\textwidth}
\centering
\def\arraystretch{1.5}
\begin{tabular}[t]{|p{0.225\textwidth}|P{0.2\textwidth}|P{0.2\textwidth}|}
 \hline 
 \backslashbox{$F$}{Noise} &   $3\%$ & $10\%$ \\
 \hline
    5 & $8.58\times10^{-3}$ & $3.53\times10^{-2}$ \\
 \hline                                                  
   15  & $8.42\times10^{-2}$ & $9.45$ \\
 \hline  
    20 & $2.47\times10$ & $2.57\times10$ \\
 \hline
\end{tabular}
\caption{{\em History-based parameterization}}
\label{tab:uq_hist}
\end{subtable}
\caption{{\em Uncertainty quantification for different sources:} (1) chaotic nature of the dynamical system, (2) parameterization limitation and (3) data quality (noise level).}
\label{tab:uncert_quan}
\end{table}

\section{Conclusion}
\label{sec:cl}

In this study, we proposed a Bayesian, history-based, parameterization for dynamical systems that defines a posterior distribution for forecasts of resolved variables that is accurate across time scales. The closure is based on a Bayesian formalism using the Hamiltonian Monte Carlo (HMC) Markov Chain sampling and also includes resolved variables states at previous time steps. This formalism does allow the definition of a stochastic parameterization.


We tested the proposed history-based parameterization on the chaotic Lorenz '96 system with noisy and sparse data. The obtained results showed its capability to produce accurate temporal forecasts for the resolved variables while returning trustworthy uncertainty quantification for different sources of error. More specifically, we proved that the proposed Bayesian history-based parameterization is capable of accurately quantifying different uncertainty sources related to (1) the chaotic and/or stochastic nature of the dynamical system, (2) the limitation introduced by the parameterization scheme compared to the ``true'' model and (3) the quality of the observational data given the noise level and temporal sparsity. Discretization of the continuous differential equations and the absence of information and data on the unresolved variables and coupling terms are other sources of uncertainty that are resolved by the Bayesian parameterization in order to return trustworthy and accurate temporal predictions for the system states. We also showed that the proposed parameterization is capable not only of learning the coupling (closure) term but also to account and correct for the numerical error introduced by the temporal discretization which enhances the stability and accuracy of the parameterized differential equation resolution.


The proposed history-based parameterization could be improved by finding a more rigorous approach to find the optimal number of previous resolved variables' states considered as inputs to the closure. Besides, other neural network architectures or different machine learning surrogate models could be investigated with the proposed history-based parameterization, such as long short-term memory neural networks. Such a modification may be critical if the proposed parameterization scheme is tested on more complex problems such as quasigeostrophic turbulence, atmospheric convection and clouds formation. Finally, the proposed parameterization scheme could be coupled with dynamical systems parameters inference approaches in order to simultaneously infer the parameterization term and the parameters values of terms with known forms. Solving the equifinality problem would be a major step in order to make such a simultaneous inference method generalizable for a variety of problems.

\section*{Competing Interests}
The authors declare that they have no competing interests.

\section*{Acknowledgement}
MAB and PG would like to thank funding from the Department of Energy DE-SC0022323 and National Science Foundation Science and Technology Center award \# 2019625 STC: Center for Learning the Earth with Artificial Intelligence and Physics (LEAP).

\section*{Author Contributions}
MAB and PG conceptualized the research. MAB designed the numerical studies and performed the simulations. PG provided funding. All authors wrote the manuscript.

\newpage
\nocite{*}
\bibliographystyle{unsrt}
\bibliography{main.bib}


\newpage
\appendix

\section{History-Based Parameterization Setup and Differentiable Time-Solver}
\label{app_sec:hist_param}

The Lorenz '96 model is defined by the following equations:
\begin{align}
    \label{eq:app_L96_X}\frac{d X_k}{dt} = -X_{k-1}(X_{k-2}-X_{k+1})-X_k+F-\frac{hc}{b}\sum\limits_{j=J(k-1)+1}^{kJ}Y_{j} \ ,k=1,\ldots,K \\
    \label{eq:app_L96_Y}\frac{d Y_{j}}{dt}=-cbY_{j+1}(Y_{j+2}-Y_{j-1})-cY_{j}+\frac{hc}{b}X_{\lfloor(j-1)/J\rfloor+1} \ , j=1,\ldots,JK \ .
\end{align}

To remedy to the ill-posed problem stemming from the inference based on parameterizations depending only on the current state of slow-varying variables, we propose a parameterized subgrid tendency that depends not only on the slow-varying variables evaluated at the current time, but also on their states at previous time-steps as follows:
\begin{equation}
    \label{eq:app_L96_param_gen}\frac{d X_k^*}{dt} = -X^*_{k-1}(X^*_{k-2}-X^*_{k+1})-X^*_k+F+P\big(X_k^*(t),X_k^*(t-\tau_1),\ldots,X_k^*(t-\tau_{n_h});\bm{\theta}\big) \ ,k=1,\ldots,K ,
\end{equation}
\noindent where $\tau_i$, $i=1,\ldots n_h$ are the time-lags that define the previous time-steps to consider for the parameterization. With such a parameterization (\ref{eq:app_L96_param_gen}), the mathematical question that should be answered in order to guarantee the well-posedness of the inference problem is: 

Given $\big\{X_k(t),X_k(t-\tau_1),\ldots,X_k(t-\tau_{n_h}), k=1\ldots,K\big\}$, are there unique $\big\{Y_j(t), j=1,\ldots JK\big\}$ such that $\big\{\big(X_k(t),Y_j(t)\big), X_k(t-\tau_1),\ldots,X_k(t-\tau_{n_h}), j=1,\ldots JK\big\}$ all belong to one trajectory that is a solution to the ``true'' model (\ref{eq:app_L96_X})-(\ref{eq:app_L96_Y})?

Providing a rigorous mathematical answer to such a question is out of the scope of the current work. However, if such a condition is satisfied, it guarantees that any future state of $(X_k, Y_j), j=1,\ldots k J, k=1\ldots,K$ is uniquely determined. Hence, an online forecast estimate $X_k^*$, $k=1,\ldots,K$ that ``matches'' the ``true'' model variables $X_k$, $k=1,\ldots,K$ can be inferred as a function of $\big\{X_k^*(t),X_k^*(t-\tau_1),\ldots,X_k^*(t-\tau_{n_h}), k=1\ldots,K\big\}$ without having access to any information on the fast-varying variables, neither on the values of the coupling term. In other words, the effect of the fast-varying variables on the forecast of the slow-varying variables is embedded within the historical (or previous multi time-steps) evolution of the slow-varying variables. Such history-based inference can be carried out using Bayesian methods and/or Machine Learning techniques as shown in the reminder of this work.


In order to explain the choice of the time-lags and time-stepping for the parameterized model, we re-write the DDE (\ref{eq:app_L96_param_gen}) as follows:
\begin{equation}
    \label{eq:app_L96_param_gen_no_k}\frac{d \bm{X}^*}{dt} = f\big(\bm{X}^*(t),\bm{X}^*(t-\tau_1),\ldots,\bm{X}^*(t-\tau_{n_h});\bm{\theta}\big) \ ,
\end{equation}
\noindent where $\bm{X}^*$ refers to the $K$-dimensional vector concatenating $X_k^*$, $k=1,\ldots,K$ and $f(\cdot)$ refers to the whole right hand-side term of equation (\ref{eq:app_L96_param_gen}).

If one chooses the time-lags as multiples of the time-step $\Delta t$ for which the data is available: $\tau_i = i \Delta t$ , $i=1,\ldots,n_h$, then applying RK4 with a time-step equal to $\Delta t$ to equation (\ref{eq:app_L96_param_gen_no_k}) gives the following time-stepping:

\begin{align}
    \label{eq:app_L96_param_gen_Deltat0}\bm{X}^*(t+\Delta t;\bm{\theta}) = \bm{X}^*(t)+\frac{1}{6}(r_1+2 \ r_2+2 \ r_3+r_4) \ , \\
   r_1 = \Delta t \ f\big(\bm{X}^*(t),\bm{X}^*(t-\Delta t),\ldots,\bm{X}^*(t-n_h \Delta t);\bm{\theta}\big) \ , \\
   r_2 = \Delta t \ f\Big(\bm{X}^*(t)+\frac{r_1}{2},\bm{X}^*(t-\frac{\Delta t}{2}),\bm{X}^*(t-\frac{3\Delta t}{2}),\ldots,\bm{X}^*\big(t-\frac{(2n_h-1)\Delta t}{2}\big);\bm{\theta}\Big) \ , \\
   r_3 = \Delta t \ f\Big(\bm{X}^*(t)+\frac{r_2}{2},\bm{X}^*(t-\frac{\Delta t}{2}),\bm{X}^*(t-\frac{3\Delta t}{2}),\ldots,\bm{X}^*\big(t-\frac{(2n_h-1)\Delta t}{2}\big);\bm{\theta}\Big) \ , \\
   \label{eq:app_L96_param_gen_Deltat4} r_4 = \Delta t \ f\Big(\bm{X}^*(t)+r_3,\bm{X}^*(t),\bm{X}^*(t-\Delta t),\ldots,\bm{X}^*\big(t-(n_h-1)\Delta t\big);\bm{\theta}\Big) \ .
\end{align}

This means that if we want to solve the parameterized model in order to fit the predicted next time-step point and fit it with the actual data-point, we would need the midpoint value of the available trajectory evaluated at a time grid with a time step equal to $\Delta t$. In order to avoid interpolating data, we actually double the time-step to integrate the parameterized model, and consider the time-lags as multiples of $2 \ \Delta t$: $\tau_i = 2 i \Delta t$ , $i=1,\ldots,n_h$. Using these settings, The RK4 time-stepping of equation (\ref{eq:app_L96_param_gen_no_k}) becomes:

\begin{align}
    \label{eq:app_L96_param_gen_2Deltat0}\bm{X}^*(t+2 \ \Delta t;\bm{\theta}) = \bm{X}^*(t)+\frac{1}{6}(r_1+2 \ r_2+2 \ r_3+r_4) \ , \\
   \label{eq:r1} r_1 = 2 \ \Delta t \ f\big(\bm{X}^*(t),\bm{X}^*(t-2 \ \Delta t),\ldots,\bm{X}^*(t-2 n_h \Delta t);\bm{\theta}\big) \ , \\
   r_2 = 2 \ \Delta t \ f\Big(\bm{X}^*(t)+\frac{r_1}{2},\bm{X}^*(t-\Delta t),\bm{X}^*(t-3 \ \Delta t),\ldots,\bm{X}^*\big(t-(2n_h-1)\Delta t\big);\bm{\theta}\Big) \ , \\
   r_3 = 2 \ \Delta t \ f\Big(\bm{X}^*(t)+\frac{r_2}{2},\bm{X}^*(t-\Delta t),\bm{X}^*(t-3 \ \Delta t),\ldots,\bm{X}^*\big(t-(2n_h-1)\Delta t\big);\bm{\theta}\Big) \ , \\
   \label{eq:app_L96_param_gen_2Deltat4} r_4 = 2 \ \Delta t \ f\Big(\bm{X}^*(t)+r_3,\bm{X}^*(t),\bm{X}^*(t-2 \Delta t),\ldots,\bm{X}^*\big(t-(2n_h-2)\Delta t\big);\bm{\theta}\Big) \ .
\end{align}

\begin{figure}
     \centering
     \begin{subfigure}[b]{0.45\textwidth}
         \centering
         \includegraphics[width=\textwidth]{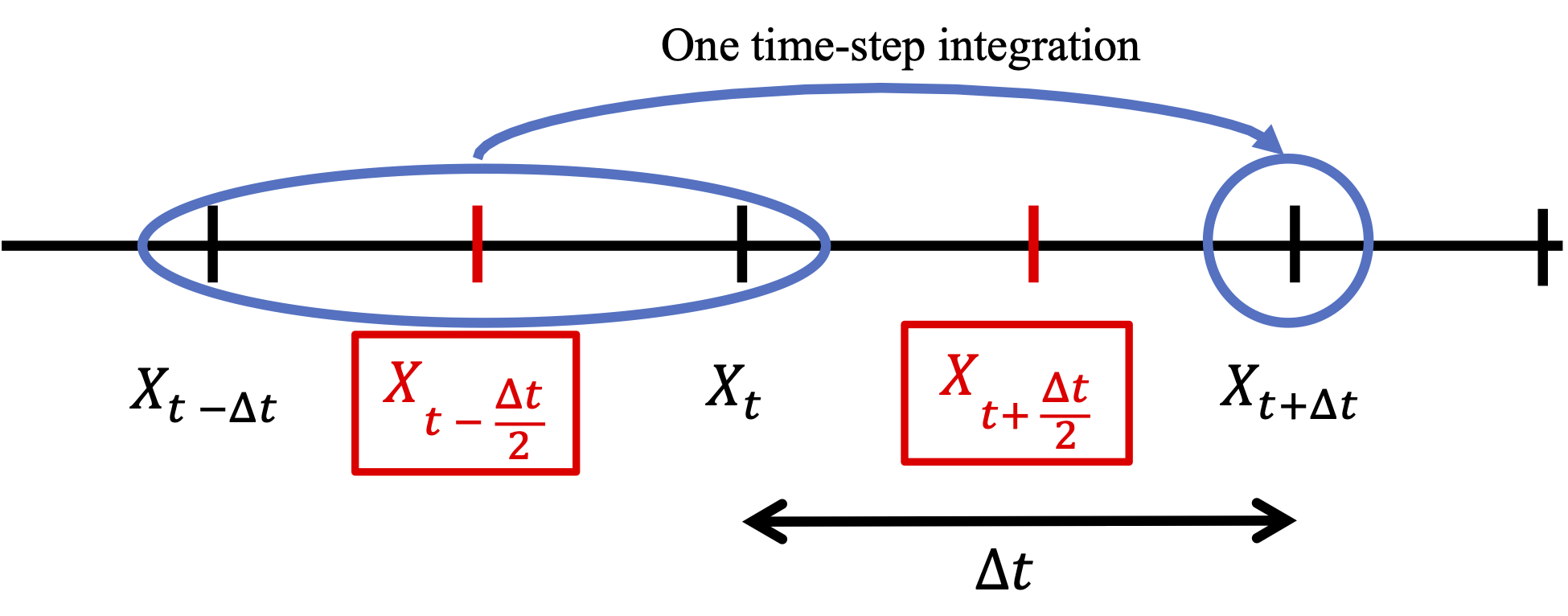}
         \caption{{\em RK4 with a time-step $\Delta t$ and time-lags as multiples of $\Delta t$ as detailed in eqs (\ref{eq:app_L96_param_gen_Deltat0})-(\ref{eq:app_L96_param_gen_Deltat4}): red points indicate the needed points for time-stepping but which are not available among the dataset.}}
         \label{fig:comput_stenc_delta_t}
     \end{subfigure}
     \begin{subfigure}[b]{0.45\textwidth}
         \centering
         \includegraphics[width=\textwidth]{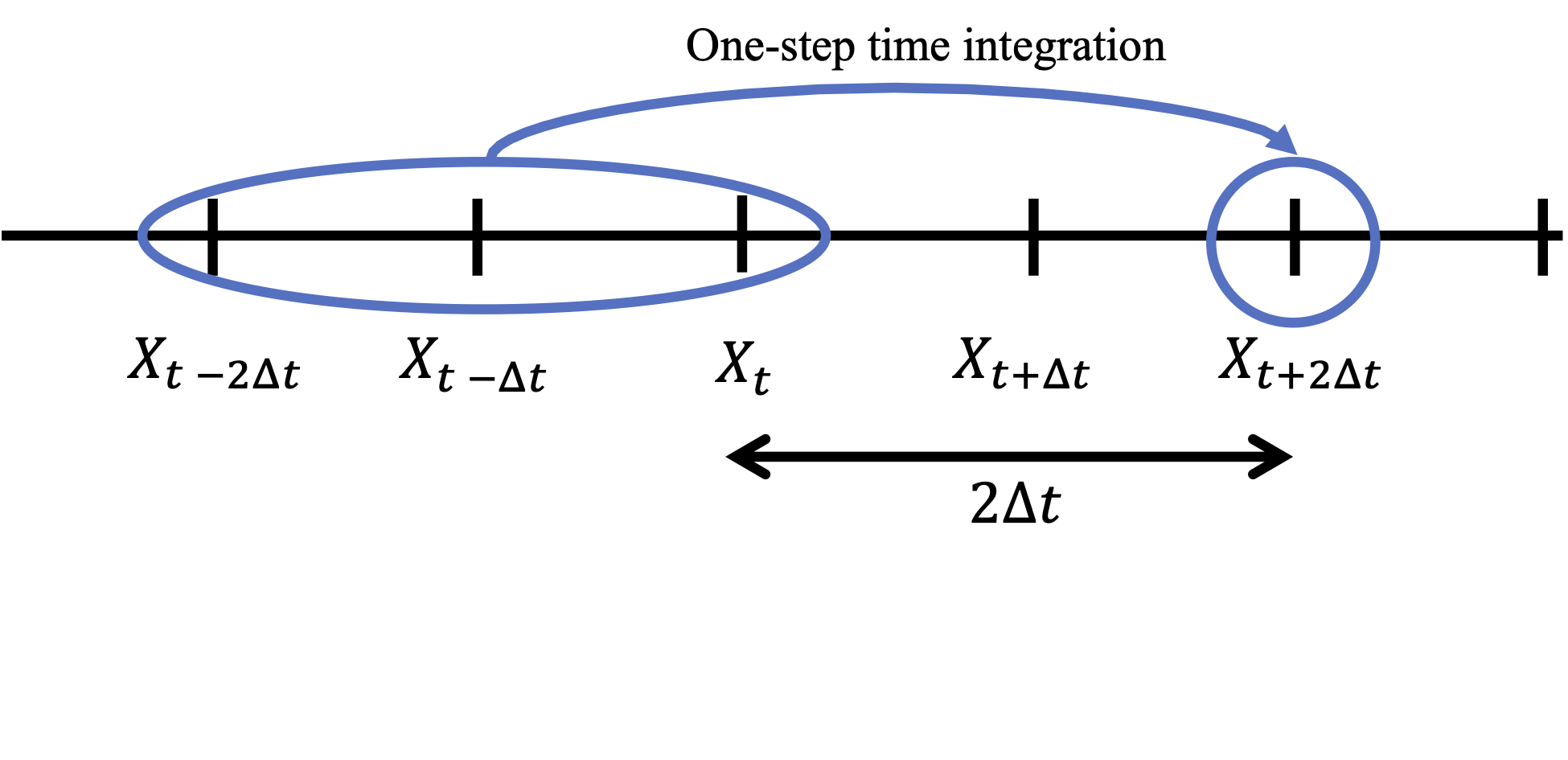}
        \caption{{\em RK4 with a time-step $2 \ \Delta t$ and time-lags as multiples of $2 \ \Delta t$ as detailed in eqs (\ref{eq:app_L96_param_gen_2Deltat0})-(\ref{eq:app_L96_param_gen_2Deltat4}).}}
         \label{fig:comput_stenc_delta_t}
     \end{subfigure}
        \caption{{\em Computational stencils for explicit RK4 applied to a DDE of the form (\ref{eq:app_L96_param_gen_no_k}): $n_h$ is taken equal to $1$ for clarity.}}
        \label{fig:comput_stenc_delta_t_and_2delta_t}
\end{figure}

This means that using the points $\{\bm{X}(t-2n_h\Delta t),\bm{X}(t-(2n_h-1)\Delta t),\ldots,\bm{X}(t-\Delta t),\bm{X}(t)\}$ from the available dataset, we can predict the point $\bm{X}^*(t+2 \ \Delta t)$ and fit the model by matching it with the data-point $\bm{X}(t+2 \ \Delta t)$. We illustrate the computational stencils associated with the time-stepping schemes (\ref{eq:app_L96_param_gen_Deltat0})-(\ref{eq:app_L96_param_gen_Deltat4}) and (\ref{eq:app_L96_param_gen_2Deltat0})-(\ref{eq:app_L96_param_gen_2Deltat4}) in figure \ref{fig:comput_stenc_delta_t_and_2delta_t}.

The time-stepping scheme detailed in (\ref{eq:app_L96_param_gen_2Deltat0})-(\ref{eq:app_L96_param_gen_2Deltat4}) provides a differentiable time-solver with respect to the unknown parameters $\bm{\theta}$. Note that in some works, an explicit second-order Runge-Kutta (RK2) scheme was considered to solve the parameterized model to represent the temporal discretization of the equations representing the resolved dynamics in an atmospheric forecasting model \cite{Gagne2020}. The numerical property detailed above for RK4 that does not require any time interpolation still applies for explicit RK2. We don't detail the corresponding algebra for sake of clarity and conciseness, as it is similar to the derivation for RK4.

\section{History-Based Parameterization Results}
\label{app:hist}

\begin{figure}
\centering
\includegraphics[width=0.8\textwidth]{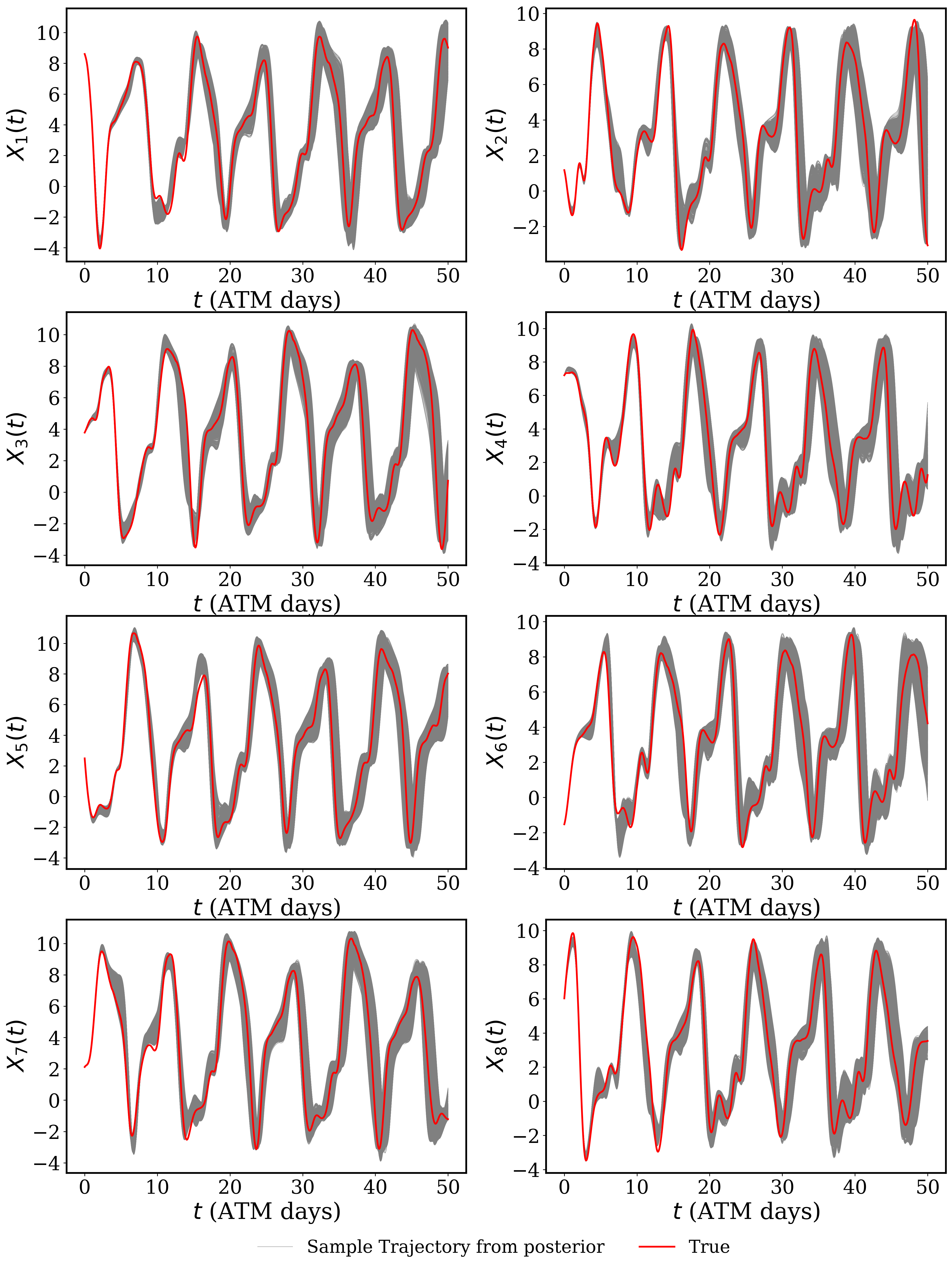}
\caption{{\em Online HMC samples predictions for the resolved slow-varying variables using the Bayesian history-based parameterization starting from the last training point:} Predictions correspond to the ensemble estimates of the parameterized DDE (\ref{eq:app_L96_param_gen}) solutions using the HMC Markov Chain parameters, while true trajectories correspond to solutions of the ``true'' model (\ref{eq:app_L96_X})-(\ref{eq:app_L96_Y}).}
\label{fig:hist_samples_X}
\end{figure}

\begin{figure}
\centering
\includegraphics[width=0.8\textwidth]{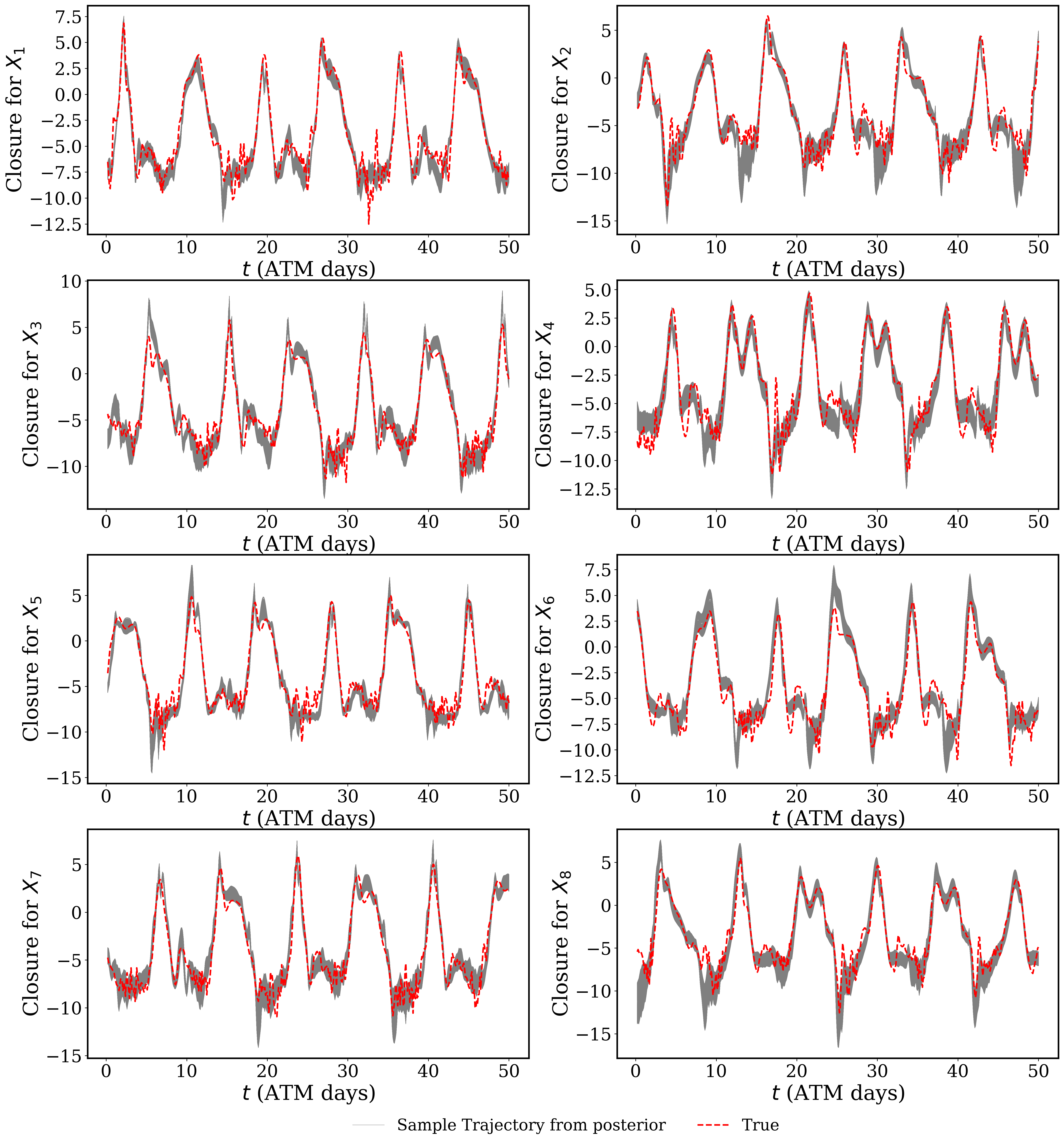}
\caption{{\em Online HMC samples predictions for the closure terms using the Bayesian history-based parameterization starting from the last training point:} Predictions correspond to the ensemble estimates of the $P(\cdot)$ terms of the parameterized DDE (\ref{eq:app_L96_param_gen}) using the HMC Markov Chain parameters, while true trajectories correspond to the coupling term in (\ref{eq:app_L96_X}).}
\label{fig:hist_samples_closure}
\end{figure}

Figures \ref{fig:hist_samples_X} and \ref{fig:hist_samples_closure} show the online HMC samples predictions for the slow-varying variables and the closure terms respectively, using the Bayesian history-based parameterization and starting from the last training point. Predictions correspond to the ensemble estimates of the parameterized DDE (\ref{eq:app_L96_param_gen}) solutions and the $P(\cdot)$ terms using the HMC Markov Chain parameters.

\section{Non-History-Based Parameterization}
\label{app:nonhist}

Existing parameterizations are designed such that they only depend on the slow-varying variables $X_k$, $k=1,\ldots K$ evaluated only at the current time. Therefore, the resulting system is expressed as an Ordinary Differential Equation (ODE):
\begin{equation}
    \label{eq:app_L96_param_other}\frac{d X_k^*}{dt} = -X^*_{k-1}(X^*_{k-2}-X^*_{k+1})-X^*_k+F+P(X_k^*;\bm{\theta}) \ ,k=1,\ldots,K ,
\end{equation}
\noindent where $X^*_k$, $k=1,\ldots,K$ is the forecast estimate of $X_k$ based on the parameterized subgrid tendency $P(\cdot;\cdot)$ and $\bm{\theta}$ is a vector of unknown parameters that are learned given the available dataset.

\begin{figure}
\centering
\includegraphics[width=0.8\textwidth]{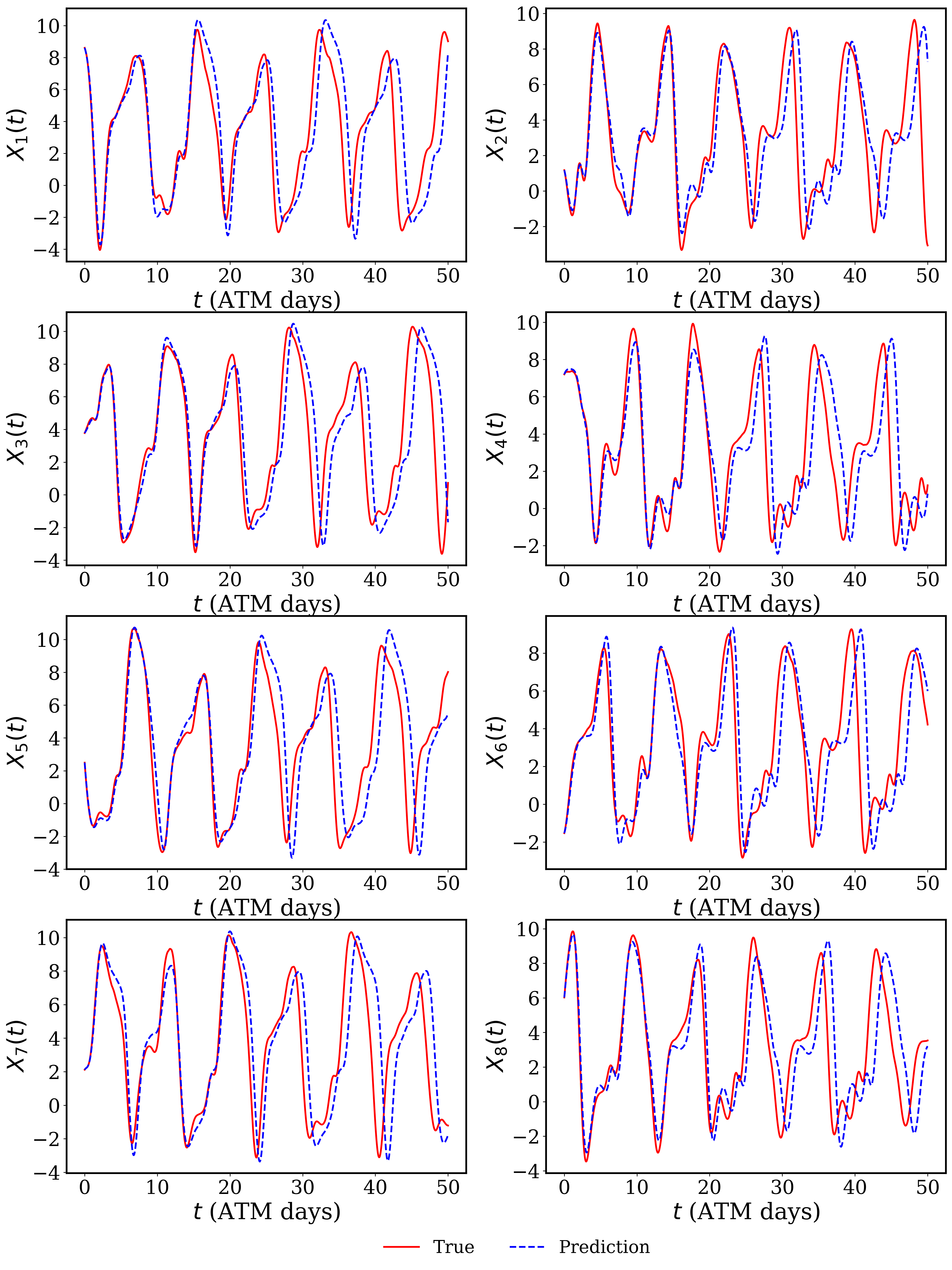}
\caption{{\em Online predictions for the resolved slow-varying variables using the deterministic non-history-based parameterization starting from the last training point:} Predictions correspond to solutions of the parameterized ODE (\ref{eq:app_L96_param_other}), while true trajectories correspond to solutions of the ``true'' model (\ref{eq:app_L96_X})-(\ref{eq:app_L96_Y}).}
\label{fig:nonhist_det_X}
\end{figure}

Figure \ref{fig:nonhist_det_X} shows the online predictions for the slow-varying variables using the deterministic non-history-based parameterization and starting from the last training point. 

\begin{figure}
\centering
\includegraphics[width=0.8\textwidth]{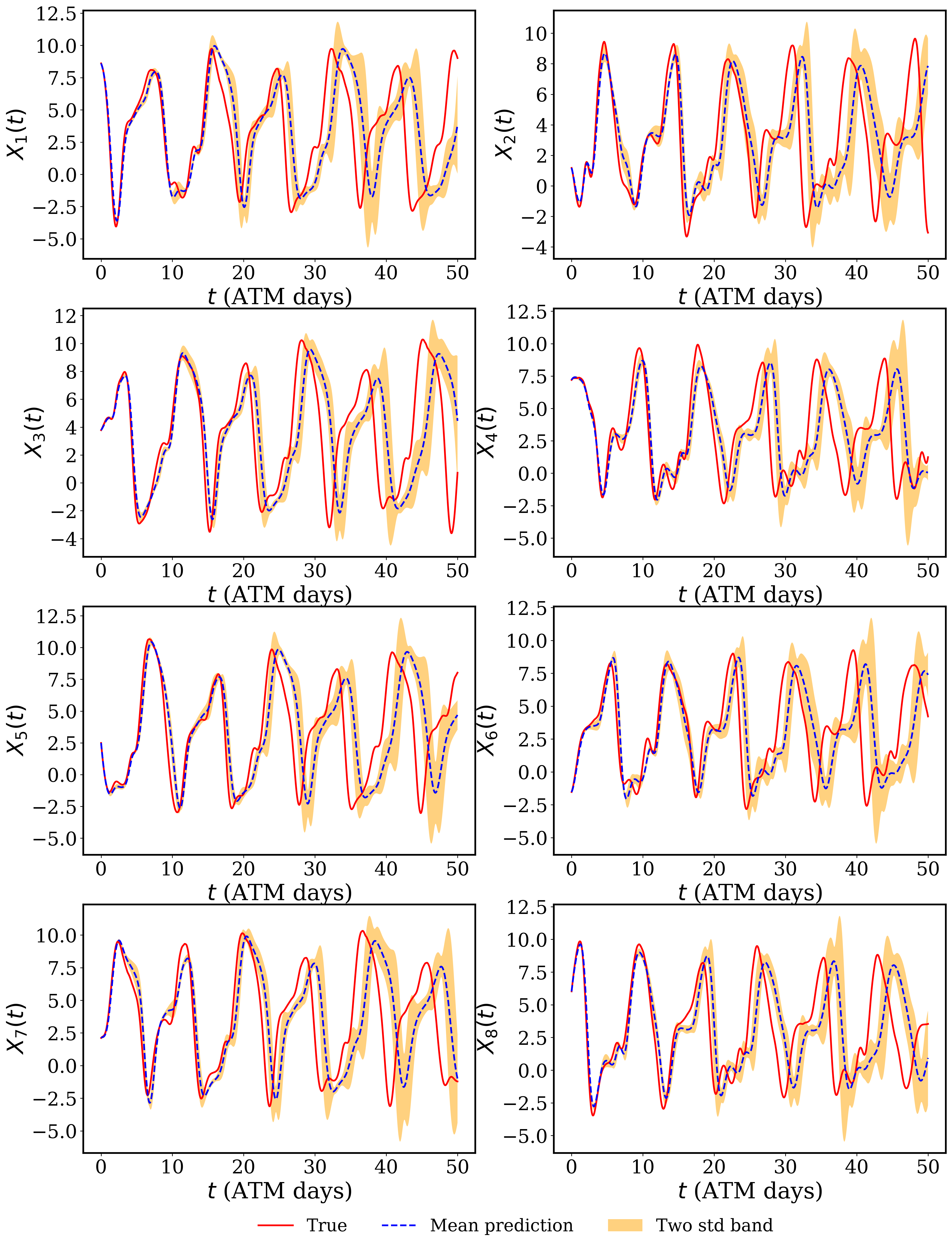}
\caption{{\em Online predictions for the resolved slow-varying variables using the Bayesian non-history-based parameterization starting from the last training point:} Predictions correspond to the ensemble estimates of the parameterized ODE (\ref{eq:app_L96_param_other}) solutions using the HMC Markov Chain parameters, while true trajectories correspond to solutions of the ``true'' model (\ref{eq:app_L96_X})-(\ref{eq:app_L96_Y}).}
\label{fig:nonhist_mean_X}
\end{figure}

\begin{figure}
\centering
\includegraphics[width=0.8\textwidth]{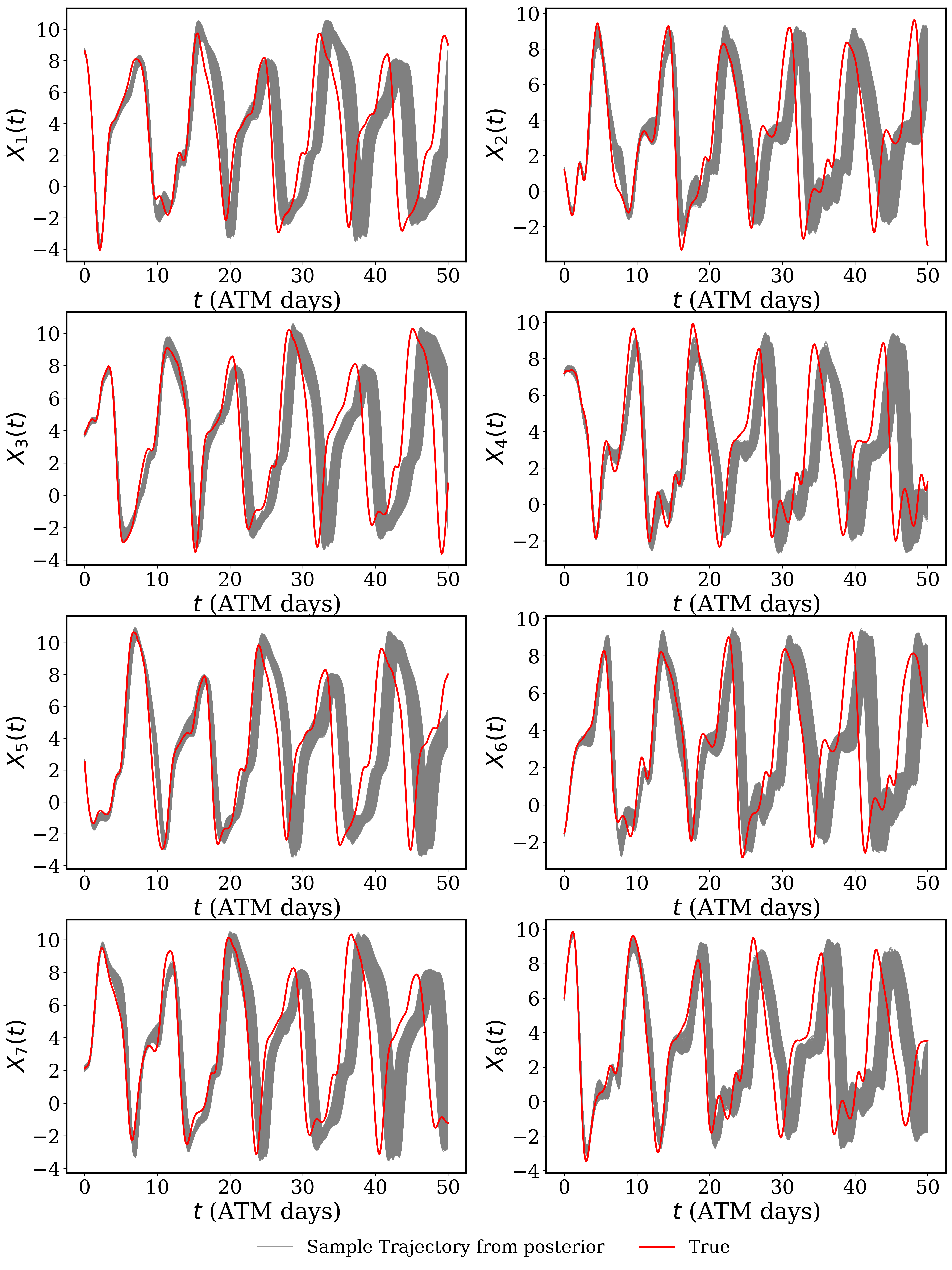}
\caption{{\em Online HMC samples predictions for the resolved slow-varying variables using the Bayesian non-history-based parameterization starting from the last training point:} Predictions correspond to the ensemble estimates of the parameterized ODE (\ref{eq:app_L96_param_other}) solutions using the HMC Markov Chain parameters, while true trajectories correspond to solutions of the ``true'' model (\ref{eq:app_L96_X})-(\ref{eq:app_L96_Y}).}
\label{fig:nonhist_samples_X}
\end{figure}

Figures \ref{fig:nonhist_mean_X} show the online predictions for the slow-varying variables using the Bayesian non-history-based parameterization and starting from the last training point and figure \ref{fig:nonhist_samples_X} shows the Bayesian predictions corresponding to the ensemble estimates of the parameterized ODE (\ref{eq:app_L96_param_other})'s solutions using the HMC Markov Chain parameters.

\end{document}